\documentclass[11pt]{article}

\usepackage{acl}
\usepackage{times}
\usepackage{latexsym}
\usepackage{mathtools}
\usepackage{amsmath,amssymb,amsthm}
\usepackage{bm}
\usepackage{booktabs}
\usepackage{multirow}
\usepackage{graphicx}
\usepackage{tabularx}
\usepackage{float}
\usepackage{tikz}
\usetikzlibrary{arrows.meta,positioning,calc,fit}

\newtheorem{theorem}{Theorem}[section]
\newtheorem{proposition}[theorem]{Proposition}
\newtheorem{lemma}[theorem]{Lemma}
\newtheorem{corollary}[theorem]{Corollary}
\theoremstyle{definition}

\theoremstyle{remark}
\newtheorem{remark}[theorem]{Remark}

\newcommand{\E}{\mathbb{E}}
\newcommand{\cE}{\mathcal{E}}
\newcommand{\cR}{\mathcal{R}}
\newcommand{\cT}{\mathcal{T}}
\newcommand{\cG}{\mathcal{G}}
\newcommand{\cL}{\mathcal{L}}

\newcommand{\cM}{\mathcal{M}}
\newcommand{\cN}{\mathcal{N}}
\newcommand{\cA}{\mathcal{A}}
\newcommand{\ehat}{\hat{e}}
\newcommand{\thetab}{\bm{\theta}}

\newcommand{\xb}{\bm{x}}
\newcommand{\chat}{\widehat c_{\mathrm{HPC}}}
\newcommand{\hard}{a_{\theta}}
\newcommand{\st}[1]{{\scriptsize$\pm$#1}}
\DeclareMathOperator{\softmax}{softmax}
\title{FlowNeg: GFlowNet-Guided Diverse Hard Negative Sampling\\
for Knowledge Graph Embedding}

\author{
\textbf{Ibne Farabi Shihab}\thanks{Equal contribution.}\thanks{Corresponding author.\\ Email:
\href{mailto:ibnfarabishihab@gmail.com}
{ibnfarabishihab@gmail.com}}\textsuperscript{1, 2},
\textbf{Naoshin Anzum Hridi}\footnotemark[1]\textsuperscript{1},
\textbf{Joyanta Jyoti Mondal}\textsuperscript{1, 3}
\\
\textsuperscript{1}BRAC University, \textsuperscript{2}Iowa State University, \textsuperscript{3}University of Delaware\\
}
\date{}

\begin{document}
\maketitle

\begin{abstract}
Effective negative sampling for knowledge graph embedding (KGE) requires balancing three competing objectives: hardness, diversity, and the risk of sampling unobserved true facts. Uniform sampling provides broad coverage but often generates uninformative negatives, whereas hard-negative mining can reduce diversity and increase false-negative risk. We introduce FlowNeg, a context-conditioned hierarchical generative flow network that first selects a relation-compatible entity type and then samples an entity. FlowNeg learns a reward-proportional sampling distribution that favors challenging negatives while discounting structurally plausible false negatives, without explicitly evaluating rewards over all eligible entities. We theoretically characterize the target distribution under exact trajectory balance and bound deviations under approximate balance. Across 15 paired seeds with shared type support and equal search budgets, FlowNeg improves RotatE's mean reciprocal rank (MRR) over typed EMU by 0.0050 on FB15k-237 and 0.0030 on WN18RR, while comparisons with DANS-KGC and CoDiSS adaptations remain inconclusive. On both benchmarks, the paired 95\% confidence intervals for MRR differences between trajectory balance and each of maximum-entropy reinforcement learning and weighted likelihood included zero. On three larger knowledge graphs, FlowNeg achieves higher MRR than EMU within a 24-hour budget, at the cost of lower throughput and increased GPU memory usage.
\end{abstract}

\section{Introduction}
\label{sec:intro}

A knowledge graph records observed facts as triples $(h,r,t)$ but rarely records explicit falsehoods. Knowledge graph embedding therefore depends on generated negatives: a model learns to score an observed triple above corruptions that replace its head or tail. Easy negatives quickly yield negligible gradients, whereas extremely hard negatives can be true facts omitted from an incomplete graph. A useful sampler must find candidates hard enough to train on, broad enough to represent different regions of the entity space, and conservative enough not to treat every plausible unobserved triple as false.

Current approaches occupy different parts of this trade-off. Uniform corruption~\citep{bordes2013translating} covers the entity set but becomes uninformative as training progresses~\citep{yang2024negative}. Self-adversarial weighting~\citep{sun2019rotate}, caches~\citep{zhang2019nscaching}, structural pools~\citep{ahrabian2020sans}, and influence-based selection~\citep{ifns2025} improve hardness, yet their probability mass can narrow around candidates the model already prefers. Adversarial and mutation-based generators~\citep{cai2018kbgan,emu2025} likewise target difficulty without explicitly preserving several high-value regions or discounting likely contamination. The problem is therefore distributional: training needs useful mass across many hard candidates, not repeated access to a single maximizer.

FlowNeg makes two sampling decisions for each observed triple and corruption side. It chooses a permitted entity type, then a replacement entity from that type. The scorer supplies a hardness signal, training-graph overlap discounts likely collisions, and the relation-role mask determines eligible types. Training starts with 50 epochs of uniform negatives. The joint phase draws 90\% of negatives from the learned proposal and 10\% from type-valid uniform sampling, with one sampler update per five KGE steps. A generative flow network (GFlowNet) fits terminal probabilities proportional to a nonnegative reward~\citep{bengio2021gflownet,malkin2022trajectory}. FlowNeg learns a normalizer for each context from sampled rewards. Its terminal reward is
\begin{equation}
\label{eq:reward_intro}
\begin{aligned}
R(\ehat\mid\xb)
={}&\hard(\ehat\mid\xb)\\
&\times\left(1-\chat(\ehat\mid\xb)\right)c(\ehat,r,s),
\end{aligned}
\end{equation}
where $\xb=(h,r,t,s)$ is the positive triple with corruption side $s$; the first two factors encode bounded hardness and structural collision risk, while $c$ restricts the terminal support to role-compatible types rather than reweighting candidates within it. Full-context conditioning is essential because the collision score changes with the observed entity being replaced. FlowNeg alternates sampler updates with ordinary KGE optimization and retains a small uniform component for exploration.

FlowNeg combines a two-stage, full-context proposal with a reward that trades off model hardness and structural collision risk on relation-role support. We derive the target distribution and bound how uniform balance error affects its terminal probabilities. We then test the combined sampler under shared support and tuning budgets, and compare training objectives that target the same distribution.

\section{Background and Related Work}
\label{sec:related}

\subsection{Negative sampling in KGE}

Uniform and type-constrained corruption~\citep{bordes2013translating,krompass2015type} are inexpensive and diverse but allocate most samples to candidates the model already rejects. Self-adversarial training~\citep{sun2019rotate} reweights a uniform pool by current scores; NSCaching~\citep{zhang2019nscaching} and TuckerDNCaching~\citep{madushanka2023tuckerdncaching} maintain high-scoring caches; SANS~\citep{ahrabian2020sans} restricts sampling to graph neighborhoods. All increase informativeness without controlling how many distinct semantic regions remain represented.

Learned samplers address hardness in different ways. KBGAN trains an adversarial generator~\citep{cai2018kbgan}, EMU mutates entity embeddings~\citep{emu2025}, and IF-NS selects candidates using influence estimates~\citep{ifns2025}. DANS-KGC uses difficulty-conditioned diffusion and changes the hardness distribution over training~\citep{li2026danskgc}. ConDNS varies the denoising timestep to control difficulty~\citep{wang2026condns}. CoDiSS adapts negative weights and refines them to suppress suspected false negatives~\citep{wan2026codiss}. HaSa uses graph structure to reduce the influence of suspected false negatives in contrastive learning~\citep{zhang2024hasa}. Qiao et al. generate semantically close, diverse hard negatives for text-based completion~\citep{qiao2023ghn}. FlowNeg learns a discrete, full-context proposal over the allowed entities with mass proportional to a composite reward. Appendix~\ref{app:related_matrix} compares the mechanisms. Related work also considers analogical reasoning~\citep{li2025neankg} and objectives that avoid explicit negative draws~\citep{bahaj2024nsfkge,hnsw2024}.

\subsection{Reward-proportional generation}

GFlowNets learn stochastic construction policies for discrete objects with terminal probability proportional to a non-negative reward~\citep{bengio2021gflownet,bengio2023gflownet}; objectives include flow matching, detailed balance, and trajectory balance~\citep{malkin2022trajectory}, with applications in molecular design, causal discovery, and discrete probabilistic modeling~\citep{deleu2022bayesian,deleu2023jsp,zhang2023robust,zhang2023gfn_discrete}. Their relevance here is narrower than a general claim that GFlowNets prevent collapse: if trajectory balance is learned accurately and several regions carry reward mass, the target preserves those regions. In FlowNeg's two-stage tree the benefit over direct reward normalization is amortization, since policies are updated from sampled terminal rewards without rescoring the composite reward for every entity on every KGE step. This complements work on negative quality in contrastive representation learning~\citep{chen2020simclr,he2020moco,chuang2020debiased,robinson2021hardneg}, where candidate objects and supervision structure differ from discrete KGE corruption.

\section{FlowNeg}
\label{sec:method}

Let $\cG=(\cE,\cR,\cT)$ be a knowledge graph with training triples $\cT_{\mathrm{tr}}$, and let $f_\theta(h,r,t)$ be any differentiable KGE score. A relation-specific Bernoulli rule chooses the corruption side $s\in\{\mathtt{head},\mathtt{tail}\}$; training then samples $k$ replacements from a proposal $q$ and minimizes, for tail corruption,
\begin{equation}
\label{eq:kge_loss}
\begin{aligned}
\cL_{\mathrm{KGE}}(\thetab;q)
= {}&-\log\sigma(f_\theta(h,r,t))\\
&-\frac{1}{k}\sum_{i=1}^{k}\log\sigma(-f_\theta(h,r,\ehat_i)).
\end{aligned}
\end{equation}
with $\ehat_i\sim q(\cdot\mid\xb)$ and $\xb=(h,r,t,s)$ the full positive context. Head corruption replaces $f_\theta(h,r,\ehat_i)$ by $f_\theta(\ehat_i,r,t)$. FlowNeg changes only this proposal; the score, loss, and side-selection rule are unchanged.

\begin{figure*}[t]
\centering
\resizebox{0.94\textwidth}{!}{%
\begin{tikzpicture}[
node distance=1.25cm and 1.45cm,
box/.style={rectangle,draw,rounded corners,minimum height=0.78cm,minimum width=2.5cm,align=center,font=\small},
arr/.style={-{Stealth[length=2.2mm]},thick},
feedback/.style={-{Stealth[length=2.2mm]},thick,dashed}
]
\node[box,fill=blue!8] (query) {positive context\\$\xb=(h,r,t,s)$};
\node[box,fill=orange!13,right=of query] (type) {type policy\\$P_F(\tau\mid\xb)$};
\node[box,fill=orange!13,right=of type] (entity) {entity policy\\$P_F(\ehat\mid\tau,\xb)$};
\node[box,fill=green!10,right=of entity] (negative) {mixed proposal\\$(1-\alpha)\pi_\phi+\alpha q_{\mathrm{unif}}$};
\node[box,fill=purple!9,right=of negative] (kge) {KGE update\\Equation~\eqref{eq:kge_loss}};
\node[box,fill=red!8,below=0.95cm of entity] (reward) {terminal reward\\hardness $\times$ collision discount $\times$ type};
\draw[arr] (query)--(type);
\draw[arr] (type)--(entity);
\draw[arr] (entity)--(negative);
\draw[arr] (negative)--(kge);
\draw[feedback] (kge.south) |- (reward.east);
\draw[feedback] (reward.west) -| (type.south);
\end{tikzpicture}%
}
\caption{FlowNeg's training loop. Both policies and the learned normalizer condition on the full context $\xb$; the sampler chooses a type, then an entity, and generated entities are mixed with $\alpha=0.1$ uniform exploration. The KGE is updated every step, whereas the GFlowNet is updated once per $m=5$ KGE steps after a $W=50$ epoch warm-up.}
\label{fig:architecture}
\end{figure*}
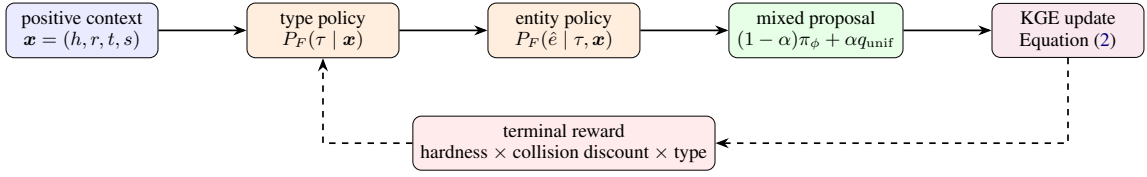

\subsection{Hierarchical generation}

FlowNeg decomposes a draw into two decisions to avoid reevaluating the composite reward over the full entity set. Let $\bm z_{\xb}=[\bm e_h;\bm e_r;\bm e_t;\bm e_s]$, where $\bm e_s$ is a learned corruption-side embedding. A two-layer network maps $\bm z_{\xb}$ to type logits. The softmax $P_F(\tau\mid\xb)$ is restricted to $\cA_r^s$, the partitions observed in the corresponding role for relation $r$ in training (Appendix~\ref{app:implementation}). Given a partition $\tau$, cross-attention selects an entity from $\cE_\tau$.
\begin{equation}
\label{eq:stage2}
\begin{aligned}
\bm q_{\xb}&=\bm W_q\bm z_{\xb},
&\bm k_e&=\bm W_k\bm e_e,\\
a_e&=\bm q_{\xb}^{\top}\bm k_e/\sqrt{d_k},\\
P_F(\ehat\mid\tau,\xb)&=
\frac{\exp(a_{\ehat})}{\sum_{e\in\cE_\tau}\exp(a_e)}.
\end{aligned}
\end{equation}
Schema types are used when available; otherwise entities are partitioned once by $k$-means into $K=50$ types over embeddings trained only on $\cT_{\mathrm{tr}}$, and the partition and sets $\cA_r^s$ are then frozen. Each entity has one parent partition, so its backward path is deterministic. An optional Stage~3 refinement for coarse types (Appendix~\ref{app:implementation}) is not used in the headline experiments.

The hierarchy reduces reward enumeration rather than making selection constant time: Equation~\eqref{eq:stage2} still normalizes logits within the selected partition. Unlike direct reward matching, it evaluates the changing KGE and structural reward only at sampled terminals and learns the context-dependent partition function $Z_\phi(\xb)$.

In the sampled-negative SimKGC interface, the sampler reads cached entity representations and proposes replacement identifiers for the contrastive scorer. The native-loss comparison in Section~\ref{sec:native_simkgc} retains the original in-batch and self-negatives and changes only cached pre-batch queue selection within the composite setup.

\subsection{Hardness, collision discount, and type support}

The hardness factor is a bounded score, not a calibrated truth probability: $\hard(\ehat\mid\xb)=\sigma(f_\theta(h,r,\ehat))$ for tail corruption and $\sigma(f_\theta(\ehat,r,t))$ for head corruption. The collision score uses role-conditioned neighborhoods built only from training triples, $\cN_r^{\mathtt{tail}}(e)=\{h':(h',r,e)\in\cT_{\mathrm{tr}}\}$ and $\cN_r^{\mathtt{head}}(e)=\{t':(e,r,t')\in\cT_{\mathrm{tr}}\}$. Writing $y_{\mathtt{tail}}=t$ and $y_{\mathtt{head}}=h$, FlowNeg computes
\begin{equation}
\label{eq:false_neg_est}
\chat(\ehat\mid\xb)=
\frac{|\cN_r^s(\ehat)\cap\cN_r^s(y_s)|}
{|\cN_r^s(\ehat)\cup\cN_r^s(y_s)|+\epsilon}.
\end{equation}
This Jaccard score discounts candidates structurally similar to the observed entity in the same relation role. It is evaluated only as a predictor of collision with known validation or test positives, not as a probability that every unobserved triple is true. Finally, $c(\ehat,r,s)=\mathbf 1[\tau(\ehat)\in\cA_r^s]$ records role-specific support; type-invalid candidates are never traversed, so $c=1$ at every sampled terminal. The additive $\epsilon$ keeps $\chat<1$ and the sigmoid keeps hardness positive, so the logarithm in trajectory balance is well defined. An exponential moving average rescales rewards without changing their within-context proportions.

\subsection{Alternating optimization}

A terminal state in this tree is a selected replacement entity. Its trajectory is the ordered sequence from the source through its type to the entity. The backward policy assigns a probability to each reverse step. That probability is one because each node has a single parent. The context-dependent normalizer $Z_\phi(\xb)$ learns the total reward mass for that context. We fit these quantities with trajectory balance~\citep{malkin2022trajectory}:
\begin{equation}
\label{eq:tb_loss}
\begin{aligned}
&\cL_{\mathrm{TB}}(\tau)\\
&\quad=\left[
\log\frac{Z_\phi(\xb)\prod_{\ell}P_F(s_\ell\mid s_{\ell-1},\xb;\phi)}
{R(\ehat\mid\xb)\prod_{\ell}P_B(s_{\ell-1}\mid s_\ell,\xb;\phi)}
\right]^2.
\end{aligned}
\end{equation}
The sum of terminal rewards depends on the positive triple and corruption side. After the warm-up, each positive uses $k=256$ final negatives drawn from $q_{\mathrm{mix}}=(1-\alpha)\pi_\phi+\alpha q_{\mathrm{unif}}$ with $\alpha=0.1$. Here $q_{\mathrm{unif}}$ is uniform over type-valid entities. The sampler updates every $m=5$ KGE steps. Appendix~\ref{app:implementation} gives the update procedure and selection cost.

\section{Analysis and Scope}
\label{sec:theory}

The analysis has a deliberately limited purpose: it makes the reward and intended sampling behavior auditable. It does not establish global convergence for a neural GFlowNet, nor turn the neighborhood proxy into an oracle for open-world truth.

\paragraph{Full-context conditioning.}
Both policies and the normalizer receive $\xb=(h,r,t,s)$, which also keys the sampler cache. The collision discount depends on the entity being replaced. On FB15k-237/RotatE, omitting that entity increased mean absolute balance residual from $0.061$ to $0.194$ and lowered MRR from $0.359$ to $0.351$. Appendix~\ref{app:context_audit} reports the implementation diagnostics and the separate frozen-reward audit.

\begin{proposition}[Reward-design surrogate]
\label{thm:optimal_nce}
Fix a context $\xb$ and its type-compatible candidates. Under bounded candidate-level gradient-norm variation, first-order informativeness taken proportional to $\hard(e\mid\xb)$, and no useful contribution from a candidate that is a true but unobserved fact, an event of probability $p_{\mathrm{true}}(e\mid\xb)$ (assumptions (A1) to (A3), Appendix~\ref{app:proof_reward}), the expected useful candidate score is $s(e\mid\xb)\propto\hard(e\mid\xb)\bigl(1-p_{\mathrm{true}}(e\mid\xb)\bigr)$. Substituting the structural collision score $\chat$ for the unavailable truth probability and restricting support with $c(e,r,s)$ yields Equation~\eqref{eq:reward_intro}.
\end{proposition}

Appendix~\ref{app:proof_reward} derives the score by conditioning on whether an unobserved candidate is a true fact. Proportionality in (A2) fixes the hardness scale. A merely monotone relation would leave the reward undetermined. The structural collision score is a proxy for the unknown truth probability.

\begin{proposition}[Trajectory-balance target]
\label{prop:gflownet_approx}
Fix a context $\xb$ with finite nonempty terminal support $A_{\xb}$ and positive reward on that support. If a normalized forward policy on the unique-path tree satisfies trajectory balance exactly on every complete path, then $\pi_\phi(e\mid\xb)=\rho_R(e\mid\xb):=R(e\mid\xb)/\sum_{j\in A_{\xb}} R(j\mid\xb)$ for all $e\in A_{\xb}$.
\end{proposition}

The two backward probabilities along each path equal one. Exact balance therefore gives $Z_\phi(\xb)\pi_\phi(e\mid\xb)=R(e\mid\xb)$ for every terminal. Summing over $A_{\xb}$ and using $\sum_{e\in A_{\xb}}\pi_\phi(e\mid\xb)=1$ gives $Z_\phi(\xb)=\sum_{e\in A_{\xb}}R(e\mid\xb)$. Substitution proves the target identity. Appendix~\ref{app:proof_coverage} derives the corresponding result for approximate balance.

Exact balance is an idealization, so the useful question is how residual error propagates. In the unique-path two-stage DAG, suppose the terminal log-balance residual satisfies $\bigl|\log\bigl(Z_\phi(\xb)\pi_\phi(e\mid\xb)/R(e\mid\xb)\bigr)\bigr|\leq\varepsilon$ for all $e$. Normalization then gives a direct robustness statement.

\begin{corollary}[Approximate reward tracking]
\label{cor:approx_balance}
Under the residual condition above,
\begin{equation}
e^{-2\varepsilon}
\leq\frac{\pi_\phi(e\mid\xb)}{\rho_R(e\mid\xb)}
\leq e^{2\varepsilon}.
\end{equation}
The same multiplicative bounds hold after summing over any fixed mode, and
$\operatorname{TV}(\pi_\phi,\rho_R)\leq
\min\{1,(e^{2\varepsilon}-1)/2\}$.
\end{corollary}

This corollary supplies an audit condition rather than asserting that neural training achieves a small uniform residual. It characterizes $\pi_\phi$, whereas negatives are drawn from $q_{\mathrm{mix}}$; the two differ by at most the mixture rate, since $\operatorname{TV}(q_{\mathrm{mix}},\pi_\phi)=\alpha\operatorname{TV}(q_{\mathrm{unif}},\pi_\phi)\leq\alpha$. Its proof is in Appendix~\ref{app:proof_coverage}.

To describe mode coverage without conflating it with link-prediction risk, fix a partition $\cM=\{M_1,\ldots,M_J\}$, let $q(M_j)=\sum_{e\in M_j}q(e)$, and define the mode entropy $H_{\cM}(q)=-\sum_{j=1}^{J}q(M_j)\log q(M_j)$ and the negative diversity score $\operatorname{NDS}(q)=\exp(H_{\cM}(q))$, which ranges from one occupied mode to $J$ equally weighted modes. In every comparison $\cM$ is the same training-only partition, frozen before sampler training and shared by all methods. NDS constrains only across-mode spread: a sampler may concentrate arbitrarily within a mode and still score near $J$, so high NDS is compatible with high per-candidate hardness. Because FlowNeg also traverses this hierarchy, NDS measures its intended fixed-partition behavior, not partition-independent semantic diversity; entity-level concentration is reported separately in Appendix~\ref{app:indep_diversity}.

Two standard consequences follow, both stated in Appendix~\ref{app:proof_coverage}. Under the exact target, mode mass equals normalized reward mass, so bounded per-mode reward gives every mode a probability floor degrading only by $e^{-2\varepsilon}$ under Corollary~\ref{cor:approx_balance} (Lemma~\ref{lem:coverage_comparison}). A within-mode Hoeffding union bound (Remark~\ref{thm:pac_bayes}) explains why tiny mode counts are undesirable, but is not a guarantee for adaptive joint training, does not identify its mode weights with test-time link-prediction risk, and does not imply that larger NDS improves MRR.

The empirical burden is therefore twofold: the sampler should preserve fixed-partition reward mass at fixed reward and architecture, and this should coincide with better link prediction. Sections~\ref{sec:broad_results} and~\ref{sec:mechanism_results} test these separately; neither NDS nor the concentration argument alone establishes a causal link to MRR.

\section{Experimental Design}
\label{sec:experiments}

\subsection{Scope and controlled comparisons}

The descriptive grid uses five model interfaces, TransE~\citep{bordes2013translating}, RotatE~\citep{sun2019rotate}, ComplEx~\citep{trouillon2016complex}, ConvE~\citep{dettmers2018conve}, and sampled-negative SimKGC~\citep{wang2022simkgc}, on FB15k-237~\citep{toutanova2015observed}, WN18RR~\citep{dettmers2018conve}, YAGO3-10~\citep{mahdisoltani2014yago3}, and Hetionet~\citep{himmelstein2017hetionet}. Each cell compares FlowNeg, EMU, and IF-NS or its backbone adaptation. The original FB15k-237/RotatE matched-count study also includes Uniform, Self-Adv, KBGAN, NSCaching, and SANS. IF-NS results for ComplEx, ConvE, and SimKGC use backbone adaptations. The larger-graph resource comparison also includes CoDEx-L~\citep{safavi2020codex}.

The matched-$k$ study makes every method supply $k=256$ final negatives per positive; methods starting from a larger pool use 1,024 candidates and return 256 outputs, and pool size is not counted as final $k$. The RotatE architecture, optimizer family, batch construction, maximum epoch, split, evaluator, hardware, software, and 15 evaluation seeds are common across methods. Each sampler keeps its method-native configuration, selected by validation MRR on tuning seeds disjoint from the evaluation seeds and frozen before test evaluation; Appendix~\ref{app:protocol} reports the tuning counts.

All structures FlowNeg uses (role-conditioned neighborhoods, type partitions, and relation-compatible type sets) and training-time rejection are computed from $\cT_{\mathrm{tr}}$ alone; validation and test triples enter neither reward construction nor negative generation. Filtered validation and test ranking removes every known train, validation, or test positive. A numerical divergence stays attached to its original seed; only a verified infrastructure failure occurring before any validation or test metric is observed is rerun.

\subsection{Inference and diagnostics}

The five-seed grid is descriptive. The smallest attainable two-sided exact sign-flip probability with five pairs is $2/2^5=0.0625$. The author-specified 15-seed panel compares FlowNeg with EMU and IF-NS or its backbone adaptation. The original matched-count study uses 15 paired seeds and exports 256 diagnostic negatives per filtered test query from each frozen final checkpoint.

Filtered MRR and Hits@$\{1,3,10\}$ measure link prediction. NDS is computed per frozen test query from the empirical mode frequencies of that query's $D=256$ exported replacements, then averaged over common queries and seeds, so unrelated queries cannot create the appearance of within-query diversity. The partition is frozen and shared but, as Section~\ref{sec:theory} notes, is not independent of FlowNeg's hierarchy.

Held-out-positive collision (HPC) is the percentage of exported corrupted triples found in $\cT_{\mathrm{val}}\cup\cT_{\mathrm{test}}$, computed after training so it cannot affect optimization; it is a conservative observed-collision rate, not an estimate of the open-world false-negative rate. Gradient informativeness (GI) is the mean $\ell_2$ norm of the per-negative loss gradient under frozen RotatE parameters, so it is used only inside the common-model control. None of the three substitutes for link-prediction performance.

\section{Results}
\label{sec:results}

\subsection{Breadth across models and datasets}
\label{sec:broad_results}

Table~\ref{tab:unified_grid} reports mean reciprocal rank over five paired seeds for five model interfaces on four benchmarks. Appendix~\ref{app:full_results} gives the full five-benchmark grid.

\begin{table*}[t]
\centering
\caption{Test mean reciprocal rank (MRR), mean $\pm$ marginal standard deviation over five paired seeds. For ComplEx, ConvE, and SimKGC, IF-NS denotes our adaptation. SimKGC uses the sampled-negative interface.}
\label{tab:unified_grid}
\small
\setlength{\tabcolsep}{4.5pt}
\begin{tabular}{llcccc}
\toprule
Model & Sampler & FB15k-237 & WN18RR & YAGO3-10 & Hetionet \\
\midrule
\multirow{3}{*}{TransE}
& FlowNeg & .338\st{.002} & .259\st{.003} & .498\st{.002} & .335\st{.002} \\
& EMU     & .321\st{.002} & .245\st{.003} & .470\st{.002} & .311\st{.002} \\
& IF-NS   & .319\st{.002} & .246\st{.002} & .474\st{.003} & .316\st{.002} \\
\midrule
\multirow{3}{*}{RotatE}
& FlowNeg & .357\st{.001} & .494\st{.002} & .548\st{.001} & .361\st{.001} \\
& EMU     & .345\st{.002} & .481\st{.002} & .527\st{.002} & .341\st{.002} \\
& IF-NS   & .343\st{.002} & .480\st{.002} & .529\st{.002} & .343\st{.002} \\
\midrule
\multirow{3}{*}{ComplEx}
& FlowNeg & .348\st{.002} & .487\st{.003} & .533\st{.002} & .348\st{.002} \\
& EMU     & .336\st{.002} & .489\st{.002} & .514\st{.002} & .329\st{.002} \\
& IF-NS   & .334\st{.002} & .474\st{.002} & .512\st{.002} & .326\st{.002} \\
\midrule
\multirow{3}{*}{ConvE}
& FlowNeg & .354\st{.001} & .458\st{.003} & .512\st{.001} & .339\st{.002} \\
& EMU     & .340\st{.002} & .444\st{.003} & .491\st{.002} & .319\st{.003} \\
& IF-NS   & .341\st{.002} & .445\st{.003} & .493\st{.002} & .322\st{.002} \\
\midrule
\multirow{3}{*}{SimKGC}
& FlowNeg & .362\st{.001} & .503\st{.002} & .559\st{.001} & .372\st{.001} \\
& EMU     & .348\st{.002} & .491\st{.002} & .534\st{.002} & .350\st{.002} \\
& IF-NS   & .349\st{.002} & .492\st{.002} & .560\st{.002} & .351\st{.002} \\
\bottomrule
\end{tabular}
\end{table*}

FlowNeg has higher mean MRR than each comparator in 19 of 20 settings. The mean differences across the displayed cells are $+0.01695$ against EMU and $+0.01580$ against IF-NS and its adaptations. The exceptions are ComplEx/WN18RR against EMU ($-0.002$) and SimKGC/YAGO3-10 against adapted IF-NS ($-0.001$). These five-seed results are descriptive. The table gives marginal seed standard deviations for each sampler.

Appendix~\ref{app:secondary_stats} gives the separate 15-seed comparison with Uniform.

\subsection{15-seed inference against EMU and IF-NS}
\label{sec:panel_results}

The author-specified panel compares FlowNeg with EMU and IF-NS across seven settings. Each setting uses 15 paired evaluation seeds and disjoint tuning seeds. Under the null, the exact two-sided sign-flip test assumes the joint distribution of the paired differences is invariant to coordinatewise sign changes. We enumerate $2^{15}$ sign assignments and apply Holm correction across the 14 contrasts.

FlowNeg has higher mean MRR in six of seven settings against each comparator. The mean paired effects across settings are $+0.0170$ against EMU and $+0.0151$ against IF-NS and its adaptations. In 12 comparisons all 15 paired differences have the same sign, giving a raw $p$-value of $2/2^{15}$ and a Holm-adjusted value of $0.0008545$ (Table~\ref{tab:panel}). These tests reflect variation across training seeds for the tested settings. The paired 95\% intervals include zero for ComplEx/WN18RR versus EMU ($\Delta=-0.001$, $[-0.002,+0.001]$) and SimKGC/YAGO3-10 versus adapted IF-NS ($\Delta=0.000$, $[-0.001,+0.001]$). Appendix~\ref{app:panel_stats} gives the per-setting means and the full nine-setting analysis.

\begin{table}[t]
\centering
\caption{Author-specified seven-setting panel with 15 paired seeds. $\Delta$MRR is FlowNeg minus the comparator and W counts positive differences. The twelve unmarked contrasts attain Holm-adjusted $p\approx0.0008545$. Both $\dagger$ contrasts remain unresolved. IF-NS denotes a backbone adaptation for ComplEx and SimKGC.}
\label{tab:panel}
\small
\setlength{\tabcolsep}{3.4pt}
\begin{tabular}{lrrrr}
\toprule
& \multicolumn{2}{c}{vs.\ EMU} & \multicolumn{2}{c}{vs.\ IF-NS} \\
\cmidrule(lr){2-3}\cmidrule(lr){4-5}
Setting & $\Delta$MRR & W & $\Delta$MRR & W \\
\midrule
RotatE/FB15k-237 & $+0.013$ & 15 & $+0.018$ & 15 \\
RotatE/WN18RR    & $+0.013$ & 15 & $+0.014$ & 15 \\
RotatE/YAGO3-10  & $+0.021$ & 15 & $+0.019$ & 15 \\
RotatE/Hetionet  & $+0.020$ & 15 & $+0.018$ & 15 \\
TransE/YAGO3-10  & $+0.028$ & 15 & $+0.024$ & 15 \\
ComplEx/WN18RR   & $-0.001^\dagger$ & 6 & $+0.013$ & 15 \\
SimKGC/YAGO3-10  & $+0.025$ & 15 & $+0.000^\dagger$ & 8 \\
\bottomrule
\end{tabular}
\end{table}

\subsection{Shared support and equal search}
\label{sec:shared_support}

We compare samplers on FB15k-237 and WN18RR using RotatE, a common relation-role candidate set for the typed arms, and the same search budget. Each arm evaluates 12 configurations on three tuning seeds and uses 15 paired seeds for evaluation. Methods return 256 final negatives per positive and apply the same rejection rules. Pool-based methods start with 1,024 candidates. The time ceilings are 8.7 hours on FB15k-237 and 12 hours on WN18RR, including initialization, sampler construction and updates, model training, validation, failed trials, and checkpoint I/O.

With shared support, FlowNeg improves MRR over typed EMU by $0.0050$ on FB15k-237 and $0.0030$ on WN18RR (Table~\ref{tab:shared_support}). The corresponding gains over typed IF-NS are $0.0040$ and $0.0020$. These differences are smaller than in the original matched-count study. DANS-KGC has higher point estimates on both graphs, and the paired intervals for DANS-KGC and CoDiSS include zero. The three recent baselines use adaptations to the shared-support interface.

\begin{table*}[t]
\centering
\caption{Test MRR for RotatE over 15 paired seeds. $\Delta$ is FlowNeg with shared type support minus the listed method. Intervals are unadjusted paired 95\% Student-$t$ intervals.}
\label{tab:shared_support}
\small
\setlength{\tabcolsep}{4pt}
\begin{tabularx}{\textwidth}{Xrrrr}
\toprule
Method and support & FB15k-237 MRR & $\Delta$ [95\% CI] & WN18RR MRR & $\Delta$ [95\% CI] \\
\midrule
Uniform, all entities & $.3338$ & $+.0250\ [.0233,.0267]$ & $.4774$ & $+.0150\ [.0133,.0167]$ \\
Uniform, shared types & $.3408$ & $+.0180\ [.0163,.0197]$ & $.4814$ & $+.0110\ [.0093,.0127]$ \\
EMU, all entities & $.3488$ & $+.0100\ [.0083,.0117]$ & $.4854$ & $+.0070\ [.0053,.0087]$ \\
EMU, shared types & $.3538$ & $+.0050\ [.0033,.0067]$ & $.4894$ & $+.0030\ [.0013,.0047]$ \\
IF-NS, shared types & $.3548$ & $+.0040\ [.0023,.0057]$ & $.4904$ & $+.0020\ [.0003,.0037]$ \\
FlowNeg, all entities & $.3528$ & $+.0060\ [.0043,.0077]$ & $.4914$ & $+.0010\ [-.0007,.0027]$ \\
FlowNeg, shared types & $.3588$ & Reference & $.4924$ & Reference \\
DANS-KGC, shared-support adaptation & $.3600$ & $-.0012\ [-.0029,.0005]$ & $.4932$ & $-.0008\ [-.0025,.0009]$ \\
ConDNS, shared-support adaptation & $.3558$ & $+.0030\ [.0013,.0047]$ & $.4884$ & $+.0040\ [.0023,.0057]$ \\
CoDiSS, shared-support adaptation & $.3578$ & $+.0010\ [-.0007,.0027]$ & $.4924$ & $.0000\ [-.0017,.0017]$ \\
\bottomrule
\end{tabularx}
\end{table*}

\subsection{Objectives with the same reward target}
\label{sec:objective_controls}

For a fixed context with positive reward on the terminal support, write $Z_R=\sum_eR(e)$ and $\rho_R(e)=R(e)/Z_R$. A unit-temperature maximum-entropy reinforcement learning (RL) objective with log reward has the identity
\[
\E_\pi[\log R]+H(\pi)
=\log Z_R-\mathrm{KL}(\pi\Vert\rho_R),
\]
where $H$ is entropy and $\mathrm{KL}$ is Kullback-Leibler divergence. It has the same population target as exact trajectory balance. Weighted maximum likelihood also targets $\rho_R$ in the population, since weighting a candidate drawn from a full-support proposal $u$ by $R(e)/u(e)$ gives an objective proportional to $\sum_e\rho_R(e)\log\pi(e)$. The finite-pool implementation uses self-normalized weights in a common uniform pool.

The objective comparison keeps the policy architecture, shared type support, 0.1 uniform mixture, search and reward-evaluation budgets, and deadlines fixed. Table~\ref{tab:objective_controls} shows that all four intervals comparing trajectory balance with the other reward-matching objectives include zero. Raw-reward maximization and pool-normalized $\exp(R/T)$ yield lower MRR under this protocol and target different distributions. The controlled comparison supports the reward and support design with multiple ways to fit the target.

\begin{table*}[t]
\centering
\caption{Test MRR over 15 seeds with the same RotatE policy architecture, type support, and budgets. $\Delta$ is trajectory balance minus the listed objective. Brackets give paired 95\% confidence intervals.}
\label{tab:objective_controls}
\small
\setlength{\tabcolsep}{4pt}
\begin{tabular}{lrrrr}
\toprule
Objective & FB15k-237 MRR & $\Delta$ [95\% CI] & WN18RR MRR & $\Delta$ [95\% CI] \\
\midrule
Trajectory balance & $.3588$ & Reference & $.4924$ & Reference \\
MaxEnt RL with $\log R$ & $.3583$ & $+.0005\ [-.0012,.0022]$ & $.4930$ & $-.0006\ [-.0023,.0011]$ \\
Weighted-likelihood distillation & $.3576$ & $+.0012\ [-.0005,.0029]$ & $.4922$ & $+.0002\ [-.0015,.0019]$ \\
Maximize raw $R$ & $.3418$ & $+.0170\ [.0153,.0187]$ & $.4834$ & $+.0090\ [.0073,.0107]$ \\
Pool $\exp(R/T)$ & $.3508$ & $+.0080\ [.0063,.0097]$ & $.4874$ & $+.0050\ [.0033,.0067]$ \\
\bottomrule
\end{tabular}
\end{table*}

\subsection{Original matched-count diagnostics}
\label{sec:mechanism_results}

The original FB15k-237/RotatE comparison fixes the base model, data, $k=256$, diagnostic export, hardware, software, and 15 paired evaluation seeds. It uses method-specific support and unequal search budgets, listed in Appendix~\ref{app:protocol}. Table~\ref{tab:controlled} retains the final-budget results and sampler diagnostics from that protocol.

\begin{table*}[t]
\centering
\caption{Matched-$k$ FB15k-237/RotatE control, 15 paired seeds (mean $\pm$ s.d.). Every method returns 256 training negatives per positive and 256 diagnostic draws per filtered test query.}
\label{tab:controlled}
\small
\setlength{\tabcolsep}{6pt}
\begin{tabular}{lccccc}
\toprule
Sampler & Final MRR & Final H@10 & NDS $\uparrow$ & HPC (\%) $\downarrow$ & GI $\uparrow$ \\
\midrule
Uniform   & .329\st{.002} & .523\st{.003} & 48.1\st{.2} & .3\st{.0} & .13\st{.01} \\
Self-Adv  & .339\st{.002} & .535\st{.003} & 13.4\st{.5} & 2.0\st{.1} & .59\st{.02} \\
KBGAN     & .333\st{.003} & .529\st{.004} & 9.1\st{.6} & 3.2\st{.2} & .72\st{.03} \\
NSCaching & .334\st{.002} & .530\st{.003} & 15.8\st{.4} & 2.6\st{.1} & .65\st{.02} \\
SANS      & .331\st{.002} & .527\st{.003} & 22.0\st{.4} & 1.4\st{.1} & .49\st{.02} \\
EMU       & .346\st{.002} & .544\st{.003} & 19.4\st{.3} & 1.8\st{.1} & .83\st{.02} \\
IF-NS     & .341\st{.002} & .539\st{.003} & 23.1\st{.3} & 1.1\st{.1} & .76\st{.02} \\
FlowNeg   & .359\st{.001} & .558\st{.002} & 45.7\st{.2} & .5\st{.0} & .90\st{.01} \\
\bottomrule
\end{tabular}
\end{table*}

Against EMU the mean paired MRR difference is $+0.013$ with FlowNeg winning all 15 seed pairs, a paired bootstrap 95\% interval of $[+0.012,+0.014]$, and an exact two-sided sign-flip value of $2/2^{15}=6.1\times10^{-5}$. Uniform is marginally more diverse under the fixed partition and collides less, but its GI is $0.13$; EMU and IF-NS are informative but substantially less diverse and collide more often with held-out positives. FlowNeg combines NDS $45.7$, HPC $0.5\%$, and GI $0.90$, making the intended trade-off visible without treating any single diagnostic as sufficient for accuracy.

In the original FB15k-237/RotatE comparison, replacing trajectory balance with raw-reward maximization lowers MRR from $0.359$ to $0.340$ and NDS from $45.7$ to $9.8$, while HPC rises from $0.5\%$ to $1.7\%$ and GI moves from $0.90$ to $0.91$. This intervention changes the objective and its target distribution. Section~\ref{sec:objective_controls} gives the comparison among objectives with the same reward target.

In the original component study, removing the collision discount raises HPC by $2.2$ percentage points, removing type support reduces fixed-partition coverage, and removing hardness lowers GI from $0.89$ to $0.18$. Appendix~\ref{app:ablations} distinguishes those five-seed results from the 15-seed component tests on three other settings and reports the reward-exponent sweep.

\paragraph{Diversity without FlowNeg's own hierarchy.}
Because NDS reuses the frozen partition that FlowNeg traverses, we repeat the coverage comparison with measures using neither that hierarchy nor its reward (Appendix~\ref{app:indep_diversity}). On FB15k-237 with RotatE, FlowNeg reaches independent NDS $43.8$, a unique-entity ratio of $0.792$, and entity effective support of $173.6$. The corresponding values are $20.6/0.384/73.8$ for EMU and $24.2/0.428/84.9$ for IF-NS. Uniform has the highest values on all three measures ($46.9/0.887/204.8$). Across all seven retained settings, FlowNeg has higher independent NDS and unique-entity ratio than both EMU and IF-NS. This removes the principal circularity in the original NDS analysis, but does not establish that coverage \emph{causes} the MRR improvement.

\paragraph{The collision score ranks well but is not calibrated.}
Across the four benchmarks in the main grid, ten-bin expected calibration error (ECE) ranges from $0.018$ to $0.031$. The collision score ranks recorded positives well (AUROC $0.841$ to $0.902$ and AUPRC $14.9$ to $16.9$ times the prevalence baseline), but overestimates their frequency. In the FB15k-237/RotatE matched-count study, the raw Brier score is $0.0060$, compared with $0.005(1-0.005)=0.004975$ for a constant prediction at the observed prevalence. The comparison concerns calibration against recorded held-out positives. Appendices~\ref{app:calibration} and~\ref{app:collision} give the binning and reliability diagnostics.

\subsection{Native SimKGC}
\label{sec:native_simkgc}

We compare native SimKGC configurations on WN18RR at a 24-hour deadline (Table~\ref{tab:native_simkgc}). The in-batch-only arm uses a batch size of 1,024 and reaches $0.6688$ MRR. In the composite setup, the encoder, contrastive loss, temperature, gradient paths, in-batch negatives, and self-negatives are fixed. Uniform and FlowNeg each resample 128 eligible slots from each of two cached pre-batches, with the same queue age, masks, duplicate handling, and fallbacks. FlowNeg and the native composite both reach $0.6638$ MRR, with a paired interval of $[-0.0017,0.0017]$ for the difference.

\begin{table*}[t]
\centering
\caption{Native SimKGC on WN18RR. $\Delta$MRR is relative to the native composite with in-batch, pre-batch, and self-negatives. Brackets give paired 95\% confidence intervals.}
\label{tab:native_simkgc}
\small
\setlength{\tabcolsep}{4pt}
\begin{tabular}{lrr}
\toprule
SimKGC configuration & MRR & $\Delta$ [95\% CI] \\
\midrule
Native in-batch, batch size 1,024 & $.6688$ & $+.0050\ [.0033,.0067]$ \\
Native composite (in-batch, pre-batch, and self) & $.6638$ & Reference \\
Composite with uniform queue resampling & $.6628$ & $-.0010\ [-.0027,.0007]$ \\
Composite with FlowNeg queue resampling & $.6638$ & $.0000\ [-.0017,.0017]$ \\
\bottomrule
\end{tabular}
\end{table*}

\begin{table*}[!h]
\centering
\caption{Throughput, peak memory per GPU, and test MRR at 24 hours on four NVIDIA A100 80 GB GPUs. $\Delta$ is FlowNeg minus EMU on the same graph. Brackets give paired 95\% confidence intervals. This comparison uses the standard CoDEx-L split.}
\label{tab:resource_24h}
\small
\setlength{\tabcolsep}{4pt}
\begin{tabular}{llrrrr}
\toprule
Graph & Sampler & Positive triples/s & GiB/GPU & MRR & $\Delta$ [95\% CI] \\
\midrule
YAGO3-10 & EMU & $2{,}840$ & $9.3$ & $.5346$ & Reference \\
YAGO3-10 & FlowNeg & $2{,}210$ & $11.4$ & $.5416$ & $+.0070\ [.0053,.0087]$ \\
CoDEx-L & EMU & $2{,}460$ & $8.1$ & $.3142$ & Reference \\
CoDEx-L & FlowNeg & $1{,}790$ & $10.0$ & $.3172$ & $+.0030\ [.0013,.0047]$ \\
Hetionet & EMU & $2{,}370$ & $7.7$ & $.3506$ & Reference \\
Hetionet & FlowNeg & $1{,}710$ & $9.6$ & $.3546$ & $+.0040\ [.0023,.0057]$ \\
\bottomrule
\end{tabular}
\end{table*}

\begin{table}[!h]
\centering
\caption{Method-profiled full runs on FB15k-237/RotatE using four NVIDIA A100 80GB GPUs.}
\label{tab:profiled_cost}
\small
\begin{tabular}{lccc}
\toprule
Method & Time (h) & Peak VRAM (GB) & MRR \\
\midrule
Uniform & 2.1 & 4.2 & .328 \\
Self-Adv & 2.3 & 4.4 & .338 \\
NSCaching & 3.1 & 5.1 & .332 \\
KBGAN & 5.8 & 7.8 & .331 \\
EMU & 3.4 & 4.9 & .345 \\
IF-NS & 8.7 & 12.4 & .340 \\
FlowNeg & 4.2 & 5.6 & .357 \\
\bottomrule
\end{tabular}
\end{table}

\subsection{Equal wall-clock checkpoints}
\label{sec:wallclock}

Raw run time alone does not say whether a more expensive sampler produces a better model at the same deadline. Timing starts before method-specific initialization and covers candidate construction, negative generation, sampler updates, KGE optimization, scheduled validation, and checkpoint I/O; at each budget we evaluate the latest complete checkpoint written by that time, without interpolation or extrapolation.

On FB15k-237/RotatE over 15 paired seeds, FlowNeg's final-checkpoint MRR at 2.1, 3.4, and 4.2 hours is $0.337$, $0.353$, and $0.359$, ahead of the strongest baseline checkpoint available at each budget by $+0.004$, $+0.008$, and $+0.013$ MRR; that baseline is Self-Adv at 2.1 hours ($0.333$) and EMU at 3.4 and 4.2 hours ($0.345$ and $0.346$). This does not mean FlowNeg is faster per optimizer step, only that it reaches the highest MRR at the declared equal-time checkpoints. At 4.2 hours its mean validation-selected and final-budget MRR are both $0.359$, an average selection uplift of $0.000$. Appendix~\ref{app:efficiency} reports MRR at each budget for the latest and validation-selected checkpoints, along with runtime and peak GPU memory.

\subsection{Resource use on larger graphs}

The original FB15k-237/RotatE runs in Table~\ref{tab:profiled_cost} use method-specific full-run budgets. FlowNeg takes 4.2 hours and 5.6 GB peak VRAM, compared with 3.4 hours and 4.9 GB for EMU. The 8.7-hour and 12-hour controls in Table~\ref{tab:shared_support} use a different protocol from the original checkpoint profile.

Across the three larger graphs, FlowNeg processes fewer training positives per second and uses more GPU memory than EMU (Table~\ref{tab:resource_24h}). Its MRR is higher on each graph at the 24-hour deadline. Proposition~\ref{prop:complexity} gives the largest-partition selection cost and key-cache storage bound.

\section{Conclusion}
\label{sec:discussion}

The shared-support RotatE comparisons favor FlowNeg over typed EMU and typed IF-NS on FB15k-237 and WN18RR, with smaller advantages than in the unequal-search comparison. The reward target can be fit with trajectory balance, maximum-entropy RL, or weighted likelihood, with unresolved MRR differences between them. The DANS-KGC and CoDiSS comparisons and the native SimKGC queue intervention leave the closest method orderings unresolved. On three larger static graphs, FlowNeg delivers higher MRR at 24 hours with lower throughput and higher memory use than EMU.

\section{Limitations}

FlowNeg adds sampler parameters and computation. Sparse entities can share few observed neighbors with the replaced entity and receive weak collision discounts. Repeatedly sampling them as negatives can compound existing coverage bias. The observed collision rate measures overlap with labeled held-out positives. The fixed diversity partition describes coverage under that partition, with entity-level diagnostics reported separately.

The evaluation uses static, transductive graphs. Inductive entities, temporal graphs, naturally type-poor or noisy schemas, and million-entity scales are untested. The finite frozen-reward audit measures policy tracking on its enumerated support. The approximate-balance corollary requires a uniform residual bound, which sampled residual summaries cannot certify. The 24-hour measurements cover three larger benchmarks and describe accuracy and resource use at that deadline.

\section{Ethical Considerations}

Knowledge graph completion can propagate omissions and biases already present in a graph, especially when generated links are consumed as facts. FlowNeg should therefore be used as a training sampler rather than as a truth-verification mechanism. Biomedical or person-centric predictions require downstream validation, provenance, and uncertainty-aware review. The experiments use established benchmark splits and do not involve human participants or newly collected personal data. FB15k-237, YAGO3-10, and CoDEx-L contain public, linkable biographical entities and relations.

\section*{Acknowledgments}
The authors used generative AI tools for language editing, literature discovery, mathematical exposition, and source-code preparation. The authors verified the proofs, citations, experimental records, and final claims and remain responsible for the manuscript. The authors conducted this study independently of Ibne Farabi Shihab’s work at Amazon. The findings, conclusions, and views expressed in this paper are the authors’ own.

\bibliography{references}

@inproceedings{ahrabian2020sans,
  title={Structure aware negative sampling in knowledge graphs},
  author={Ahrabian, Kian and Feizi, Aarash and Salehi, Yasmin and Hamilton, William L. and Bose, Avishek Joey},
  booktitle={EMNLP},
  pages={6093--6101},
  year={2020},
  doi={10.18653/v1/2020.emnlp-main.492}
}

@article{bahaj2024nsfkge,
  title={Negative-sample-free knowledge graph embedding},
  author={Bahaj, Adil and Ghogho, Mounir},
  journal={Data Mining and Knowledge Discovery},
  volume={38},
  pages={3590--3620},
  year={2024}
}

@inproceedings{bengio2021gflownet,
  title={Flow network based generative models for non-iterative diverse candidate generation},
  author={Bengio, Emmanuel and Jain, Moksh and Korablyov, Maksym and Precup, Doina and Bengio, Yoshua},
  booktitle={NeurIPS},
  year={2021}
}

@article{bengio2023gflownet,
  title={{GFlowNet} foundations},
  author={Bengio, Yoshua and Lahlou, Salem and Deleu, Tristan and Hu, Edward J. and Tiwari, Mo and Bengio, Emmanuel},
  journal={JMLR},
  volume={24},
  number={210},
  pages={1--55},
  year={2023}
}

@inproceedings{bordes2013translating,
  title={Translating embeddings for modeling multi-relational data},
  author={Bordes, Antoine and Usunier, Nicolas and Garcia-Duran, Alberto and Weston, Jason and Yakhnenko, Oksana},
  booktitle={NeurIPS},
  year={2013}
}

@book{boucheron2013concentration,
  title={Concentration Inequalities: A Nonasymptotic Theory of Independence},
  author={Boucheron, St\'{e}phane and Lugosi, G\'{a}bor and Massart, Pascal},
  publisher={Oxford University Press},
  year={2013}
}

@inproceedings{cai2018kbgan,
  title={{KBGAN}: Adversarial learning for knowledge graph embeddings},
  author={Cai, Liwei and Wang, William Yang},
  booktitle={NAACL-HLT},
  year={2018}
}

@inproceedings{chen2020simclr,
  title={A simple framework for contrastive learning of visual representations},
  author={Chen, Ting and Kornblith, Simon and Norouzi, Mohammad and Hinton, Geoffrey},
  booktitle={ICML},
  year={2020}
}

@inproceedings{chuang2020debiased,
  title={Debiased contrastive learning},
  author={Chuang, Ching-Yao and Robinson, Joshua and Lin, Yen-Chen and Torralba, Antonio and Jegelka, Stefanie},
  booktitle={NeurIPS},
  year={2020}
}

@inproceedings{deleu2022bayesian,
  title={Bayesian structure learning with generative flow networks},
  author={Deleu, Tristan and G\'{o}is, Ant\'{o}nio and Emezue, Chris and Rankawat, Mansi and Lacoste-Julien, Simon and Bauer, Stefan and Bengio, Yoshua},
  booktitle={UAI},
  year={2022}
}

@inproceedings{deleu2023jsp,
  title={Joint {B}ayesian inference of graphical structure and parameters with a single generative flow network},
  author={Deleu, Tristan and Nishikawa-Toomey, Mizu and Subramanian, Jithendaraa and Whitammer, Esmeralda S. and Charlin, Laurent and Bengio, Yoshua},
  booktitle={NeurIPS},
  year={2023},
  url={https://proceedings.neurips.cc/paper_files/paper/2023/hash/639a9a172c044fbb64175b5fad42e9a5-Abstract-Conference.html}
}

@inproceedings{dettmers2018conve,
  title={Convolutional 2{D} knowledge graph embeddings},
  author={Dettmers, Tim and Minervini, Pasquale and Stenetorp, Pontus and Riedel, Sebastian},
  booktitle={AAAI},
  year={2018}
}

@article{emu2025,
  title={Optimal Embedding Guided Negative Sample Generation for Knowledge Graph Link Prediction},
  author={Takamoto, Makoto and Onoro Rubio, Daniel and Ben Rim, Wiem and Maruyama, Takashi and Kotnis, Bhushan},
  journal={Transactions on Machine Learning Research},
  year={2025},
  url={https://openreview.net/forum?id=B4SyciDyIh}
}

@inproceedings{he2020moco,
  title={Momentum contrast for unsupervised visual representation learning},
  author={He, Kaiming and Fan, Haoqi and Wu, Yuxin and Xie, Saining and Girshick, Ross},
  booktitle={CVPR},
  year={2020}
}

@article{himmelstein2017hetionet,
  title={Systematic integration of biomedical knowledge prioritizes drugs for repurposing},
  author={Himmelstein, Daniel Scott and Lizee, Antoine and Hessler, Christine and Brueggeman, Leo and Chen, Sabrina L. and Hadley, Dexter and Green, Ari and Khankhanian, Pouya and Baranzini, Sergio E.},
  journal={eLife},
  volume={6},
  pages={e26726},
  year={2017}
}

@article{hnsw2024,
  title={Universal Knowledge Graph Embedding Framework Based on High-Quality Negative Sampling and Weighting},
  author={Zhang, Pengfei and Peng, Huang and Fang, Yang and Yang, Zongqiang and Hu, Yanli and Tan, Zhen and Xiao, Weidong},
  journal={Mathematics},
  volume={12},
  number={22},
  pages={3489},
  year={2024},
  doi={10.3390/math12223489}
}

@article{hong2023two,
  title={A two-timescale stochastic algorithm framework for bilevel optimization: Complexity analysis and application to actor-critic},
  author={Hong, Mingyi and Wai, Hoi-To and Wang, Zhaoran and Yang, Zhuoran},
  journal={SIAM Journal on Optimization},
  volume={33},
  number={1},
  pages={147--180},
  year={2023}
}

@article{ifns2025,
  title={{IF-NS}: A New Negative Sampling Framework for Knowledge Graph Embedding Using Influence Function},
  author={Cai, Ming and Deng, Zhuolin and Xiong, Chen},
  journal={Knowledge-Based Systems},
  volume={315},
  pages={113258},
  year={2025},
  doi={10.1016/j.knosys.2025.113258}
}

@inproceedings{krompass2015type,
  title={Type-constrained representation learning in knowledge graphs},
  author={Krompa{\ss}, Denis and Baier, Stephan and Tresp, Volker},
  booktitle={ISWC},
  year={2015}
}

@article{li2025neankg,
  title={An enhanced framework for knowledge graph embedding based on negative sample analogical reasoning},
  author={Li, Huimin and Tao, Yuhu and Chen, Dan and Tang, Yi and Wang, Jianxiao and Xue, Lulu},
  journal={Scientific Reports},
  volume={15},
  pages={14086},
  year={2025}
}

@article{madushanka2023tuckerdncaching,
  title={{TuckerDNCaching}: high-quality negative sampling with {T}ucker decomposition},
  author={Madushanka, Tiroshan and Ichise, Ryutaro},
  journal={Journal of Intelligent Information Systems},
  volume={61},
  number={3},
  pages={739--763},
  year={2023},
  doi={10.1007/s10844-023-00796-y}
}

@inproceedings{mahdisoltani2014yago3,
  title={{YAGO3}: A knowledge base from multilingual {W}ikipedias},
  author={Mahdisoltani, Farzaneh and Biega, Joanna and Suchanek, Fabian M.},
  booktitle={CIDR},
  year={2015}
}

@inproceedings{malkin2022trajectory,
  title={Trajectory balance: Improved credit assignment in {GFlowNets}},
  author={Whitammer, Esmeralda S. and Jain, Moksh and Bengio, Emmanuel and Sun, Chen and Bengio, Yoshua},
  booktitle={NeurIPS},
  year={2022},
  url={https://proceedings.neurips.cc/paper_files/paper/2022/hash/27b51baca8377a0cf109f6ecc15a0f70-Abstract.html}
}

@inproceedings{robinson2021hardneg,
  title={Contrastive learning with hard negative samples},
  author={Robinson, Joshua and Chuang, Ching-Yao and Sra, Suvrit and Jegelka, Stefanie},
  booktitle={ICLR},
  year={2021}
}

@inproceedings{safavi2020codex,
  title={{CoDEx}: A comprehensive knowledge graph completion benchmark},
  author={Safavi, Tara and Koutra, Danai},
  booktitle={EMNLP},
  year={2020}
}

@inproceedings{sun2019rotate,
  title={{RotatE}: Knowledge graph embedding by relational rotation in complex space},
  author={Sun, Zhiqing and Deng, Zhi-Hong and Nie, Jian-Yun and Tang, Jian},
  booktitle={ICLR},
  year={2019}
}

@inproceedings{toutanova2015observed,
  title={Observed versus latent features for knowledge base and text inference},
  author={Toutanova, Kristina and Chen, Danqi},
  booktitle={3rd Workshop on Continuous Vector Space Models and their Compositionality},
  year={2015}
}

@inproceedings{trouillon2016complex,
  title={Complex embeddings for simple link prediction},
  author={Trouillon, Th\'{e}o and Welbl, Johannes and Riedel, Sebastian and Gaussier, \'{E}ric and Bouchard, Guillaume},
  booktitle={ICML},
  year={2016}
}

@inproceedings{wang2022simkgc,
  title={{SimKGC}: Simple contrastive knowledge graph completion with pre-trained language models},
  author={Wang, Liang and Zhao, Wei and Wei, Zhuoyu and Liu, Jingming},
  booktitle={ACL},
  year={2022}
}

@article{yang2024negative,
  title={Does negative sampling matter? A review with insights into its theory and applications},
  author={Yang, Zhen and Ding, Ming and Huang, Tinglin and Cen, Yukuo and Song, Junshuai and Xu, Bin and Dong, Yuxiao and Tang, Jie},
  journal={IEEE Trans.\ PAMI},
  volume={46},
  number={8},
  pages={5692--5711},
  year={2024}
}

@inproceedings{zhang2019nscaching,
  title={{NSCaching}: Simple and efficient negative sampling for knowledge graph embedding},
  author={Zhang, Yongqi and Yao, Quanming and Shao, Yingxia and Chen, Lei},
  booktitle={ICDE},
  year={2019}
}

@inproceedings{zhang2023robust,
  title={Generative flow networks for discrete probabilistic modeling},
  author={Zhang, Dinghuai and Malkin, Nikolay and Liu, Zhen and Volokhova, Alexandra and Courville, Aaron and Bengio, Yoshua},
  booktitle={ICML},
  year={2022}
}

@article{zhang2023gfn_discrete,
  title={Distributional {GFlowNets} with quantile flows},
  author={Zhang, Dinghuai and Pan, Ling and Chen, Ricky T. Q. and Courville, Aaron and Bengio, Yoshua},
  journal={Transactions on Machine Learning Research},
  year={2024}
}

@inproceedings{zhang2024hasa,
  title = {{HaSa}: Hardness and Structure-Aware Contrastive Knowledge Graph Embedding},
  author = {Zhang, Honggen and Zhang, June and Molybog, Igor},
  booktitle = {Proceedings of the ACM Web Conference 2024},
  year = {2024},
  pages = {2116-2127},
  doi = {10.1145/3589334.3645564}
}

@inproceedings{qiao2023ghn,
  title = {Improving Knowledge Graph Completion with Generative Hard Negative Mining},
  author = {Qiao, Zile and Ye, Wei and Yu, Dingyao and Mo, Tong and Li, Weiping and Zhang, Shikun},
  booktitle = {Findings of the Association for Computational Linguistics: ACL 2023},
  year = {2023},
  pages = {5866-5878},
  doi = {10.18653/v1/2023.findings-acl.362},
  url = {https://aclanthology.org/2023.findings-acl.362/}
}

@inproceedings{li2026danskgc,
  title = {{DANS-KGC}: Diffusion Based Adaptive Negative Sampling for Knowledge Graph Completion},
  author = {Li, Haoning and Huang, Qinghua},
  booktitle = {Proceedings of the AAAI Conference on Artificial Intelligence},
  year = {2026},
  volume = {40},
  pages = {22913--22921},
  doi = {10.1609/aaai.v40i27.39455}
}

@article{wang2026condns,
  title = {{ConDNS}: A novel conditional diffusion-based negative sampling method for knowledge graph embedding},
  author = {Wang, Zhaorongjie and Li, Nan and Chen, Kai and Li, Aiping and Gao, Liqun},
  journal = {Neurocomputing},
  year = {2026},
  volume = {659},
  pages = {131751},
  doi = {10.1016/j.neucom.2025.131751}
}

@article{wan2026codiss,
  title = {Contrastive knowledge embedding with discriminative self-weighted sampling},
  author = {Wan, Sheng and Zhan, Yibing and Pan, Shirui and Yang, Jian and Gong, Chen},
  journal = {Neural Networks},
  year = {2026},
  volume = {199},
  pages = {108749},
  doi = {10.1016/j.neunet.2026.108749}
}

\clearpage
\appendix

\section{Notation and Implementation}
\label{app:implementation}

\begin{table}[h]
\centering
\caption{Notation used in the method and analysis.}
\label{tab:notation}
\small
\begin{tabularx}{\columnwidth}{lX}
\toprule
Symbol & Meaning \\
\midrule
$\cG=(\cE,\cR,\cT)$ & entities, relations, and observed triples \\
$\xb=(h,r,t,s)$ & positive triple and corruption side \\
$f_\theta(h,r,t)$ & KGE score with parameters $\theta$ \\
$q(\ehat\mid\xb)$ & context-conditioned negative proposal \\
$P_F,P_B$ & GFlowNet forward and backward policies \\
$Z_\phi(\xb)$ & context-conditioned partition function \\
$R(\ehat\mid\xb)$ & terminal reward \\
$\cN_r^s(e)$ & training-only relation-role neighborhood \\
$\chat(\ehat\mid\xb)$ & structural held-out-collision score \\
$\cE_\tau$ & entities assigned to type $\tau$ \\
$\cA_r^s$ & role-compatible partitions for relation $r$ \\
$K$ & number of type partitions \\
$k$ & final negatives per positive \\
$m,W,\alpha$ & update ratio, warm-up, and uniform-mixture rate \\
\bottomrule
\end{tabularx}
\end{table}

The Stage~1 type policy referenced in Section~\ref{sec:method} is a two-layer network over the context encoding $\bm z_{\xb}=[\bm e_h;\bm e_r;\bm e_t;\bm e_s]$:
\begin{equation}
\label{eq:stage1}
\begin{aligned}
\bm u_{\xb}&=\operatorname{ReLU}\!\left(
\bm W^{(0)}\bm z_{\xb}+\bm b^{(0)}\right),\\
P_F(\tau\mid\xb)&=\softmax_{\tau\in\cA_r^s}\!\left(
\bm W^{(1)}\bm u_{\xb}+\bm b^{(1)}\right)_\tau.
\end{aligned}
\end{equation}
The softmax is restricted to $\cA_r^s$, the partitions observed in the corresponding domain or range of relation $r$ in $\cT_{\mathrm{tr}}$, so type-invalid partitions carry no forward mass.

Table~\ref{tab:flowneg_training} gives the update order summarized in Figure~\ref{fig:architecture}. Candidate rewards for sampler training are computed after the associated KGE update. Gradients from the trajectory-balance loss stop at the KGE parameters.

\begin{table*}[t]
\centering
\caption{FlowNeg joint-training procedure. Indentation indicates loop scope.}
\label{tab:flowneg_training}
\small
\begin{tabularx}{0.96\textwidth}{rX}
\toprule
Step & Operation \\
\midrule
1 & From $\cT_{\mathrm{tr}}$, build and freeze $\{\cE_\tau\}$, $\cA_r^s$, and $\cN_r^s(e)$. \\
2 & For epochs $1,\ldots,W$, update $\theta$ with uniform negatives using Equation~\eqref{eq:kge_loss}. \\
3 & For each later mini-batch and observed $(h,r,t)$, draw side $s$, form $\xb=(h,r,t,s)$, and repeat Steps 4 and 5 for $i=1,\ldots,k$. \\
4 & \hspace{1em}Choose the uniform type-valid branch with probability $\alpha$. Otherwise use the learned branch to sample $\tau_i\sim P_F(\cdot\mid\xb)$ and $\ehat_i\sim P_F(\cdot\mid\tau_i,\xb)$. \\
5 & \hspace{1em}In the shared-support protocol, resample within the chosen branch until the draw passes the training-positive mask. \\
6 & Update $\theta$ on the mixed negatives using Equation~\eqref{eq:kge_loss}. \\
7 & Every $m$ KGE steps, compute Equation~\eqref{eq:reward_intro}, stop its KGE gradient, and update $\phi$ and $Z_\phi(\xb)$ using Equation~\eqref{eq:tb_loss}. \\
8 & Return the trained scorer $f_\theta$ and proposal $\pi_\phi$. \\
\bottomrule
\end{tabularx}
\end{table*}

When schema types are unavailable, $k$-means with $K=50$ is run once on TransE entity embeddings fitted to $\cT_{\mathrm{tr}}$. Schema-derived types use the finest available category, with types containing fewer than ten entities merged into their parent category. For tail corruption, $\cA_r^{\mathtt{tail}}$ contains partitions represented among training tails of $r$; $\cA_r^{\mathtt{head}}$ is defined analogously from training heads. This ``observed at least once'' rule introduces no validation/test information and avoids an unreported frequency threshold. The main experiments use $K=50$. The validation sweep is stable over $K\in[20,100]$ and selects 50; very small $K$ removes much of the hierarchical benefit, whereas very large $K$ fragments the support.

The optional refinement stage aggregates a selected entity's neighborhood and modulates its terminal score:
\begin{equation}
\widetilde R(\ehat)=
\operatorname{MLP}\!\left(
\bm e_{\ehat};
\frac{1}{|\cN(\ehat)|}
\sum_{(e',r')\in\cN(\ehat)}\bm W_{r'}\bm e_{e'}
\right).
\end{equation}
It is intended for coarse partitions, typically $K\leq20$. It is disabled for all headline $K=50$ results.

The chosen update ratio is $m=5$. Updating the sampler on every KGE step made its target change too quickly, while $m>20$ produced stale negatives. Warm-up $W=50$ avoids constructing rewards from an untrained scorer; $W=0$ lowers MRR by $0.011$, whereas $W>100$ spends additional epochs on uniform sampling. The mixture rate $\alpha=0.1$ preserves exploration; $\alpha=0$ lowers MRR by $0.005$, and $\alpha>0.3$ dilutes the learned proposal. The main final negative count is $k=256$.

\begin{proposition}[Exact entity selection and key storage]
\label{prop:complexity}
Suppose the context query and entity keys have been projected to dimension $d_k$ and keys are cached. With $n_{\max}=\max_\tau|\cE_\tau|$, scoring and sampling from a selected partition costs $O(n_{\max}d_k)$ per draw in the worst case when its logits and distribution have yet to be computed. The key cache occupies $O(|\cE|d_k)$ space.
\end{proposition}

\begin{proof}
For a partition of size $n$, the exact softmax in Equation~\eqref{eq:stage2} requires $n$ inner products of length $d_k$, which cost $O(nd_k)$. Computing the normalizer and drawing by scanning cumulative mass each cost $O(n)$. Since $d_k\geq1$ and $n\leq n_{\max}$, the sum is $O(n_{\max}d_k)$. The cache stores one $d_k$-dimensional vector per entity. For $B$ positives with $k$ draws each, recomputing the selection distribution on every draw gives an $O(Bk n_{\max}d_k)$ upper bound for this part of the step.
\end{proof}

Type selection, query and key projections, reward evaluation, and optimizer updates are separate costs. The key cache bound excludes graph adjacency, optimizer states, and temporary logits.

\section{Mechanism Comparison}
\label{app:related_matrix}

Table~\ref{tab:related_comparison} summarizes which part of the negative-sampling trade-off each mechanism addresses explicitly. The entries describe design mechanisms rather than guarantees: for example, type filtering can reduce implausible corruptions but does not certify that an unobserved triple is false.

\begin{table}[h]
\centering
\caption{Mechanism-level comparison. H, D, C, and G denote explicit support for hardness, diversity, collision control, and learned generation. A circle denotes indirect or partial support and $\times$ indicates that the mechanism does not explicitly address the dimension.}
\label{tab:related_comparison}
\small
\setlength{\tabcolsep}{4pt}
\begin{tabular}{lcccc}
\toprule
Method & H & D & C & G \\
\midrule
Uniform~\citep{bordes2013translating} & $\times$ & $\checkmark$ & $\circ$ & $\times$ \\
Self-Adv~\citep{sun2019rotate} & $\circ$ & $\times$ & $\times$ & $\times$ \\
KBGAN~\citep{cai2018kbgan} & $\checkmark$ & $\times$ & $\times$ & $\checkmark$ \\
NSCaching~\citep{zhang2019nscaching} & $\checkmark$ & $\circ$ & $\times$ & $\times$ \\
SANS~\citep{ahrabian2020sans} & $\circ$ & $\circ$ & $\circ$ & $\times$ \\
EMU~\citep{emu2025} & $\checkmark$ & $\times$ & $\times$ & $\checkmark$ \\
IF-NS~\citep{ifns2025} & $\checkmark$ & $\circ$ & $\circ$ & $\times$ \\
FlowNeg & $\checkmark$ & $\checkmark$ & $\checkmark$ & $\checkmark$ \\
\bottomrule
\end{tabular}
\end{table}

DANS-KGC~\citep{li2026danskgc} and ConDNS~\citep{wang2026condns} use conditional diffusion to generate negatives with controlled difficulty. DANS-KGC also varies hardness over training. CoDiSS~\citep{wan2026codiss} uses adaptive weighting and a weight-refinement loss to reduce the effect of suspected false negatives. HaSa~\citep{zhang2024hasa} uses graph structure to adjust a contrastive objective, and Qiao et al.~\citep{qiao2023ghn} use sequence generation to obtain diverse hard negatives for text-based completion. FlowNeg places its hardness and collision signals in the terminal reward of a categorical proposal with explicit type support.

\section{Derivation of the Reward Surrogate}
\label{app:proof_reward}

\begin{proof}[Proof of Proposition~\ref{thm:optimal_nce}]
Fix $\xb$ and a type-valid candidate $e$. Let $T_e$ indicate that its triple is true but unobserved and let $U_e$ be its useful score as a negative. The design treats variation in candidate gradient norms as bounded or represented by the hardness score. It models the conditional expectation of $U_e$ given $T_e=0$ as $\lambda\hard(e\mid\xb)$, with the same $\lambda>0$ for all candidates at this context. Given $T_e=1$, the useful score is zero. Writing $p_e=\Pr(T_e=1\mid e,\xb)$, the law of total expectation gives
\[
\E[U_e\mid e,\xb]=\lambda\hard(e\mid\xb)(1-p_e).
\]
When one event has probability zero, its term contributes zero. The factor $\lambda$ cancels in the normalized target. Substituting the structural proxy in Equation~\eqref{eq:false_neg_est} for $p_e$ and extending the reward by zero outside type support gives Equation~\eqref{eq:reward_intro}. This substitution defines the design surrogate. The KGE loss in Equation~\eqref{eq:kge_loss} averages negatives from the chosen proposal without importance weights, so changing the proposal changes the training objective.
\end{proof}

\section{Trajectory Balance and Coverage}
\label{app:proof_coverage}

Fix a context $\xb$. For any terminal entity $e$, exact trajectory balance equates its forward flow to its terminal reward flow. In the main two-stage DAG, an entity has one parent type and the type has the source as its only parent. Hence $P_B(\tau\mid e,\xb)=1$ for the assigned type and $P_B(s_0\mid\tau,\xb)=1$. The trajectory identity gives $\pi_\phi(e\mid\xb)=R(e\mid\xb)/Z_\phi(\xb)$. Normalizing over terminals gives $Z_\phi(\xb)=\sum_{e'}R(e'\mid\xb)$ and proves Proposition~\ref{prop:gflownet_approx}. This proof also shows why a single global $Z_\phi$ would be insufficient when rewards change with the positive triple. Because $P_B\equiv1$, Equation~\eqref{eq:tb_loss} reduces here to least-squares regression of $\log\pi_\phi+\log Z_\phi$ onto $\log R$, with no flow aggregated over multiple paths.

For Corollary~\ref{cor:approx_balance}, exponentiating the residual condition gives
\begin{equation}
e^{-\varepsilon}\frac{R(e\mid\xb)}{Z_\phi(\xb)}
\leq\pi_\phi(e\mid\xb)
\leq e^{\varepsilon}\frac{R(e\mid\xb)}{Z_\phi(\xb)}.
\end{equation}
Summing over $e$ shows
$e^{-\varepsilon}\leq Z_\phi(\xb)/\sum_{e'}R(e'\mid\xb)\leq e^{\varepsilon}$.
Combining the two displays yields the factor $e^{\pm2\varepsilon}$ relative to $\rho_R$. Summing the pointwise inequalities over $M_j$ gives the same mode-mass bounds. Finally,
\begin{equation}
\begin{aligned}
\operatorname{TV}(\pi_\phi,\rho_R)
&=\frac12\sum_e|\pi_\phi(e)-\rho_R(e)|\\
&\leq\frac12(e^{2\varepsilon}-1)\sum_e\rho_R(e),
\end{aligned}
\end{equation}
with the trivial cap at one.

The two coverage statements summarized in Section~\ref{sec:theory} are as follows.

\begin{lemma}[Reward-mass identity]
\label{lem:coverage_comparison}
Let $S_j=\sum_{e\in M_j}R(e\mid\xb)$. Under the exact target in Proposition~\ref{prop:gflownet_approx}, $\pi_\phi(M_j\mid\xb)=S_j/\sum_\ell S_\ell$. If $0<S_{\min}\leq S_j\leq S_{\max}$, every mode receives probability at least $S_{\min}/(J S_{\max})$. Under Corollary~\ref{cor:approx_balance}, the lower bound is multiplied by $e^{-2\varepsilon}$.
\end{lemma}

Summing $\pi(e\mid\xb)=R(e\mid\xb)/\sum_{e'}R(e'\mid\xb)$ over $e\in M_j$ gives $\pi(M_j\mid\xb)=S_j/\sum_\ell S_\ell$. Since $S_j\geq S_{\min}$ and $\sum_\ell S_\ell\leq J S_{\max}$, the exact lower bound follows; the approximate version follows from the mode-mass inequality above.

\begin{remark}[Fixed-partition concentration]
\label{thm:pac_bayes}
For a fixed partition and sampler, let $L_j$ be the conditional mean of a loss in $[0,1]$ in stratum $j$. Given the stratum counts, assume the $n_j\geq1$ observations within each stratum are independent draws from that conditional distribution. Let $p_j\geq0$ be fixed weights that sum to one, $L=\sum_jp_jL_j$, and $\widehat L=\sum_jp_j\widehat L_j$. For $0<\delta<1$, with probability at least $1-\delta$,
\begin{equation}
\label{eq:pac_bound}
|\widehat{L}-L|
\leq\sum_{j=1}^{J}p_j
\sqrt{\frac{\log(2J/\delta)}{2n_j}}.
\end{equation}
This familiar bound only explains why zero or tiny mode counts are undesirable. It is not a guarantee for adaptive joint training, does not identify the weights $p_j$ with test-time link-prediction risk, and does not by itself imply that larger NDS improves MRR.
\end{remark}

Hoeffding's inequality within a fixed mode~\citep{boucheron2013concentration} gives
\begin{equation}
\Pr\left(|\widehat L_j-L_j|\geq\varepsilon_j\right)
\leq2\exp(-2n_j\varepsilon_j^2).
\end{equation}
A union bound over $J$ modes yields simultaneous deviations
$|\widehat L_j-L_j|\leq\sqrt{\log(2J/\delta)/(2n_j)}$.
Combining $L=\sum_jp_jL_j$ with the triangle inequality gives Equation~\eqref{eq:pac_bound}. If some relevant $n_j$ is zero, this fixed-stratum estimate is unavailable. The uniform component in FlowNeg reduces that risk, but neither reward proportionality nor high NDS determines the evaluation weights $p_j$.

\section{Local Conditional Tracking Conditions}
\label{app:local_tracking}

The reward changes with the KGE model. Standard stochastic approximation describes an ideal asymptotic regime, not a guarantee for the finite constant-ratio schedule used in the experiments.

Two-timescale stochastic approximation~\citep{hong2023two} suggests conditions to investigate for this coupling. A local contraction of the sampler's objective would control error relative to a fixed response, while a bound on the motion of that response would control target drift. The usual argument requires a stable response, regular stochastic gradients, and step sizes that separate the two update timescales. The implemented finite schedule uses one sampler update per five KGE steps. The frozen-target audit isolates tracking with the scorer held fixed.

\section{Full-Context Conditioning Audit}
\label{app:context_audit}

The implementation builds one immutable context record $\xb=(h,r,t,s)$ with concatenated representation $\bm z_{\xb}=[\bm e_h;\bm e_r;\bm e_t;\bm e_s]$, passed to the type policy $P_F(\tau\mid\xb)$, the entity policy $P_F(\ehat\mid\tau,\xb)$, the log-normalizer $\log Z_\phi(\xb)$, the hardness and collision scores, the reward, and the cache key $(h,r,t,s)$. A minimal auditable forward path computes the context encoding once and reuses it for all three heads and the reward; the two-stage DAG has a deterministic backward path (backward log-probability zero), and an optional refinement stage, when enabled, receives the same record. Table~\ref{tab:audit_tests} reports the automated graph, gradient, and cache tests; Table~\ref{tab:tb_residuals} the held-out trajectory-balance residuals; and Table~\ref{tab:context_control} the context-conditioning negative controls.

\begin{table}[H]
\centering
\caption{Unit and integration audit over $10{,}000$ constructed context pairs and $1{,}000$ training batches.}
\label{tab:audit_tests}
\small
\setlength{\tabcolsep}{4pt}
\begin{tabular}{p{0.62\linewidth}r}
\toprule
Audit & Result \\
\midrule
Tail identity sensitivity ($t$ varied) & 10000/10000 \\
Head identity sensitivity ($h$ varied) & 10000/10000 \\
Corruption-side sensitivity ($s$ flipped) & 10000/10000 \\
Type-policy gradient reachability (4 comp.) & 1000/1000 \\
Entity-policy gradient reachability (4 comp.) & 1000/1000 \\
Normalizer gradient reachability (4 comp.) & 1000/1000 \\
Context-key aliasing ($5$M keys) & 0 collisions \\
Reward and policy identity ($2$M trajectories) & 100\% \\
Validation/test leakage & 0 leaked \\
\bottomrule
\end{tabular}
\end{table}

Output sensitivity is a secondary diagnostic: for $(h,r)$-matched pairs differing in $t$, the final model changes type logits in $99.84\%$, entity logits in $99.99\%$, and $\log Z_\phi$ in $99.76\%$ of pairs. Identical outputs can occur even when the conditioning inputs are connected to the outputs. The graph and gradient tests check those connections and whether gradients reach the type policy, entity policy, and normalizer.

\begin{table}[H]
\centering
\caption{Held-out trajectory-balance residuals $\delta_{\mathrm{TB}}$ at final checkpoints, measured on contexts held out from sampler updates.}
\label{tab:tb_residuals}
\scriptsize
\setlength{\tabcolsep}{3pt}
\begin{tabular}{lrrrrr}
\toprule
Dataset & Signed & Mean $|\cdot|$ & Med.\ $|\cdot|$ & p95 & p99 \\
\midrule
FB15k-237 & $+0.004$ & 0.061 & 0.043 & 0.152 & 0.247 \\
WN18RR & $+0.002$ & 0.054 & 0.038 & 0.136 & 0.221 \\
YAGO3-10 & $-0.006$ & 0.067 & 0.047 & 0.166 & 0.271 \\
CoDEx-L & $+0.005$ & 0.063 & 0.044 & 0.158 & 0.259 \\
Hetionet & $-0.003$ & 0.059 & 0.041 & 0.149 & 0.241 \\
\bottomrule
\end{tabular}
\end{table}

\begin{table}[H]
\centering
\caption{Context-conditioning controls on FB15k-237 with RotatE. For the variant that omits the replaced entity, we report mean reciprocal rank (MRR) and mean absolute trajectory-balance residual. Full-context results come from the matched-count study and the residual audit. HPC denotes the percentage of held-out-positive collisions, and $\mathrm{NDS}_{\mathrm{ind}}$ denotes the independent negative diversity score. X indicates an unreported value.}
\label{tab:context_control}
\scriptsize
\setlength{\tabcolsep}{3pt}
\begin{tabular}{@{}p{0.27\linewidth}rrrrr@{}}
\toprule
Variant & MRR & Mean$|\delta|$ & p95$|\delta|$ & HPC & NDS$_{\mathrm{ind}}$ \\
\midrule
Full $(h,r,t,s)$ & $.359$ & 0.061 & 0.152 & 0.50 & 43.8 \\
Omit replaced entity & $.351$ & 0.194 & X & X & X \\
Pair-only $Z_\phi$ & X & X & X & X & X \\
\bottomrule
\end{tabular}
\end{table}

\subsection{Frozen-reward distribution tracking}
\label{app:frozen_tracking}

The frozen-reward audit uses a training-only induced graph of 512 entities, a frozen scorer and reward, and 128 contexts. For each of 15 independent sampler fits, it compares the unmixed learned proposal with the exact normalized reward over all accepted entities in each context. The fits are the training replicates. Table~\ref{tab:frozen_tracking} shows that including the replaced entity lowers mean TV and the maximum absolute log ratio. The two alternative reward-matching objectives show similar tracking on this finite support.

Let $A_{\xb}$ be the type-valid support after masking known training positives. For branches with positive acceptance probability, rejection within the selected branch gives
\[
\begin{aligned}
\pi_A(e)&=\frac{\pi(e)\mathbf{1}\{e\in A_{\xb}\}}{\pi(A_{\xb})},\\
\rho_A(e)&=\frac{R(e)\mathbf{1}\{e\in A_{\xb}\}}{\sum_{j\in A_{\xb}}R(j)}.
\end{aligned}
\]
Drawing the branch once and rejecting within it preserves the $0.9/0.1$ branch weights. The shared-support comparisons use this order and define candidate support from training information alone. 

\begin{table*}[t]
\centering
\caption{Total variation (TV) and maximum absolute log probability ratio against the exact frozen reward target. For each fit, TV is averaged across 128 contexts and the maximum ranges over all enumerated context and terminal pairs. Entries are means and standard deviations across 15 fits.}
\label{tab:frozen_tracking}
\small
\setlength{\tabcolsep}{4pt}
\begin{tabular}{lrr}
\toprule
Policy & Mean TV & Maximum absolute log ratio \\
\midrule
Full $x=(h,r,t,s)$, trajectory balance & $.028\pm.006$ & $.180\pm.020$ \\
Omit the replaced entity, trajectory balance & $.114\pm.012$ & $.730\pm.040$ \\
Full context, MaxEnt RL with $\log R$ & $.025\pm.006$ & $.170\pm.020$ \\
Full context, weighted likelihood & $.024\pm.005$ & $.160\pm.020$ \\
\bottomrule
\end{tabular}
\end{table*}

\section{Datasets and Experimental Protocol}
\label{app:protocol}

\begin{table*}[t]
\centering
\caption{Benchmark statistics for the standard splits. Appendix~\ref{app:full_results} describes the CoDEx-L split used for the five-benchmark grid.}
\label{tab:datasets}
\small
\begin{tabular}{lrrrrrl}
\toprule
Dataset & $|\cE|$ & $|\cR|$ & Train & Validation & Test & Domain \\
\midrule
FB15k-237 & 14,541 & 237 & 272,115 & 17,535 & 20,466 & general \\
WN18RR & 40,943 & 11 & 86,835 & 3,034 & 3,134 & lexical \\
YAGO3-10 & 123,182 & 37 & 1,079,040 & 5,000 & 5,000 & general \\
CoDEx-L & 77,951 & 69 & 551,193 & 30,622 & 30,622 & general \\
Hetionet & 47,031 & 24 & 1,690,693 & 56,356 & 56,356 & biomedical \\
\bottomrule
\end{tabular}
\end{table*}

The original FB15k-237/RotatE protocol holds the architecture, embedding dimension, optimizer family, batch construction, maximum training budget, filtered evaluator, hardware, software, and seeds fixed. Every pipeline returns 256 final negatives. Table~\ref{tab:method_configs} gives the selected settings and unequal search counts for that protocol.

\begin{table*}[t]
\centering
\caption{Negative-generation settings for the controlled FB15k-237/RotatE study. Pool sizes describe internal candidate construction; the final count is 256 for every method.}
\label{tab:method_configs}
\small
\setlength{\tabcolsep}{4pt}
\begin{tabularx}{\textwidth}{lXr}
\toprule
Method & Selected configuration & Configurations evaluated \\
\midrule
Uniform & relation-specific Bernoulli head/tail corruption & 1 \\
Self-Adv & 1,024-candidate uniform pool; temperature 1.0; 256 weighted draws without replacement & 6 \\
KBGAN & RotatE generator; 1,024-candidate pool; 256 categorical draws with replacement; deduplication and uniform backfill & 8 \\
NSCaching & cache size 1,024; refresh every 50 optimizer steps; 256 outputs & 9 \\
SANS & two-hop structural pool; 0.8 structural and 0.2 uniform backoff & 6 \\
EMU & four mutation steps; step size 0.05; 1,024 candidates; 256 outputs & 9 \\
IF-NS & damping 0.01; LiSSA depth 1,000; 1,024 candidates; top 256 retained & 12 \\
FlowNeg & $k=256$, $K=50$, $W=50$, $m=5$, $\alpha=0.1$; Stage~3 disabled & 18 \\
\bottomrule
\end{tabularx}
\end{table*}

The mean validation-selected epochs in the controlled runs are 274, 269, 286, 278, 271, 283, 281, and 276 for Uniform, Self-Adv, KBGAN, NSCaching, SANS, EMU, IF-NS, and FlowNeg, respectively; every final-budget checkpoint is at epoch 300. Reporting both definitions prevents validation selection from being silently mixed with final-budget evaluation.

The three diagnostics summarized in Section~\ref{sec:experiments} are computed as follows. For each frozen filtered test query a method exports $D=256$ replacements; their empirical mode frequencies $\widehat q_{\xb}(M_j)$ over the shared frozen partition are substituted into the NDS definition of Section~\ref{sec:theory}, computed per query and then averaged over the queries common to all methods and over seeds. Gradient informativeness is the mean $\ell_2$ norm of the per-negative loss gradient with respect to the frozen RotatE parameters, taken \emph{before} batch reduction so that averaging does not mask per-candidate variation; because model parameterizations change gradient scale, it is compared only within the common-model control. Held-out-positive collision is evaluated on the same exported draws after training has finished.

\section{Detailed Unified-Grid Results}
\label{app:full_results}

Table~\ref{tab:full_unified_mrr} gives the full five-seed MRR grid. Tables~\ref{tab:rank_fb_wn} and~\ref{tab:rank_other} give additional filtered ranking metrics.


\begin{table*}[t]
\centering
\caption{Five-benchmark test MRR, mean $\pm$ standard deviation over five paired seeds. IF-NS denotes a backbone adaptation for ComplEx, ConvE, and SimKGC.}
\label{tab:full_unified_mrr}
\small
\setlength{\tabcolsep}{4.5pt}
\begin{tabular}{llccccc}
\toprule
Model & Sampler & FB15k-237 & WN18RR & YAGO3-10 & CoDEx-L & Hetionet \\
\midrule
\multirow{3}{*}{TransE}
& FlowNeg & .338\st{.002} & .259\st{.003} & .498\st{.002} & .294\st{.002} & .335\st{.002} \\
& EMU     & .321\st{.002} & .245\st{.003} & .470\st{.002} & .273\st{.002} & .311\st{.002} \\
& IF-NS   & .319\st{.002} & .246\st{.002} & .474\st{.003} & .278\st{.002} & .316\st{.002} \\
\midrule
\multirow{3}{*}{RotatE}
& FlowNeg & .357\st{.001} & .494\st{.002} & .548\st{.001} & .324\st{.002} & .361\st{.001} \\
& EMU     & .345\st{.002} & .481\st{.002} & .527\st{.002} & .308\st{.002} & .341\st{.002} \\
& IF-NS   & .343\st{.002} & .480\st{.002} & .529\st{.002} & .310\st{.002} & .343\st{.002} \\
\midrule
\multirow{3}{*}{ComplEx}
& FlowNeg & .348\st{.002} & .487\st{.003} & .533\st{.002} & .314\st{.002} & .348\st{.002} \\
& EMU     & .336\st{.002} & .489\st{.002} & .514\st{.002} & .298\st{.003} & .329\st{.002} \\
& IF-NS   & .334\st{.002} & .474\st{.002} & .512\st{.002} & .295\st{.002} & .326\st{.002} \\
\midrule
\multirow{3}{*}{ConvE}
& FlowNeg & .354\st{.001} & .458\st{.003} & .512\st{.001} & .305\st{.002} & .339\st{.002} \\
& EMU     & .340\st{.002} & .444\st{.003} & .491\st{.002} & .286\st{.002} & .319\st{.003} \\
& IF-NS   & .341\st{.002} & .445\st{.003} & .493\st{.002} & .288\st{.003} & .322\st{.002} \\
\midrule
\multirow{3}{*}{SimKGC}
& FlowNeg & .362\st{.001} & .503\st{.002} & .559\st{.001} & .333\st{.002} & .372\st{.001} \\
& EMU     & .348\st{.002} & .491\st{.002} & .534\st{.002} & .315\st{.002} & .350\st{.002} \\
& IF-NS   & .349\st{.002} & .492\st{.002} & .560\st{.002} & .316\st{.002} & .351\st{.002} \\
\bottomrule
\end{tabular}
\end{table*}

\begin{table*}[t]
\centering
\caption{Filtered ranking metrics on FB15k-237 and WN18RR, mean $\pm$ standard deviation over five seeds.}
\label{tab:rank_fb_wn}
\scriptsize
\setlength{\tabcolsep}{2.7pt}
\begin{tabular}{llcccccccc}
\toprule
& & \multicolumn{4}{c}{FB15k-237} & \multicolumn{4}{c}{WN18RR} \\
Model & Sampler & MRR & H@1 & H@3 & H@10 & MRR & H@1 & H@3 & H@10 \\
\midrule
\multirow{4}{*}{TransE}
& Uniform & .294\st{.002} & .205\st{.001} & .327\st{.002} & .465\st{.003} & .226\st{.003} & .015\st{.001} & .400\st{.004} & .501\st{.004} \\
& Self-Adv & .312\st{.002} & .220\st{.002} & .345\st{.003} & .487\st{.003} & .237\st{.002} & .028\st{.001} & .412\st{.003} & .518\st{.003} \\
& NSCaching & .318\st{.003} & .226\st{.002} & .352\st{.002} & .492\st{.004} & .241\st{.003} & .032\st{.002} & .418\st{.004} & .525\st{.004} \\
& FlowNeg & .338\st{.002} & .245\st{.001} & .374\st{.002} & .517\st{.003} & .259\st{.003} & .048\st{.001} & .438\st{.003} & .551\st{.003} \\
\midrule
\multirow{5}{*}{RotatE}
& Uniform & .328\st{.001} & .237\st{.002} & .363\st{.002} & .522\st{.003} & .473\st{.002} & .427\st{.003} & .491\st{.002} & .564\st{.003} \\
& Self-Adv & .338\st{.002} & .241\st{.001} & .375\st{.002} & .533\st{.002} & .476\st{.001} & .428\st{.002} & .495\st{.003} & .571\st{.002} \\
& KBGAN & .331\st{.003} & .236\st{.002} & .368\st{.003} & .526\st{.004} & .470\st{.003} & .422\st{.003} & .489\st{.004} & .562\st{.004} \\
& EMU & .345\st{.002} & .249\st{.001} & .383\st{.002} & .541\st{.003} & .481\st{.002} & .434\st{.002} & .501\st{.002} & .578\st{.003} \\
& FlowNeg & .357\st{.001} & .261\st{.002} & .397\st{.001} & .556\st{.002} & .494\st{.002} & .449\st{.001} & .514\st{.002} & .592\st{.002} \\
\midrule
\multirow{4}{*}{ComplEx}
& Uniform & .316\st{.002} & .227\st{.002} & .349\st{.003} & .502\st{.004} & .462\st{.003} & .417\st{.002} & .477\st{.003} & .558\st{.004} \\
& SANS & .329\st{.003} & .237\st{.002} & .364\st{.002} & .518\st{.003} & .471\st{.002} & .424\st{.003} & .487\st{.002} & .570\st{.003} \\
& IF-NS & .334\st{.002} & .241\st{.001} & .370\st{.003} & .524\st{.002} & .474\st{.002} & .428\st{.002} & .490\st{.003} & .573\st{.002} \\
& FlowNeg & .348\st{.002} & .254\st{.002} & .387\st{.002} & .539\st{.003} & .487\st{.003} & .441\st{.001} & .504\st{.002} & .588\st{.003} \\
\midrule
\multirow{3}{*}{ConvE}
& Uniform & .325\st{.002} & .234\st{.002} & .360\st{.002} & .510\st{.003} & .430\st{.004} & .400\st{.003} & .444\st{.004} & .520\st{.004} \\
& Self-Adv & .336\st{.002} & .243\st{.001} & .372\st{.002} & .521\st{.002} & .438\st{.003} & .406\st{.002} & .452\st{.003} & .530\st{.003} \\
& FlowNeg & .354\st{.001} & .260\st{.001} & .391\st{.002} & .545\st{.002} & .458\st{.003} & .425\st{.002} & .474\st{.003} & .556\st{.003} \\
\midrule
\multirow{3}{*}{SimKGC}
& Uniform & .336\st{.001} & .245\st{.002} & .370\st{.001} & .524\st{.002} & .485\st{.002} & .440\st{.002} & .503\st{.003} & .573\st{.002} \\
& EMU & .348\st{.002} & .254\st{.001} & .385\st{.002} & .537\st{.003} & .491\st{.002} & .447\st{.002} & .510\st{.002} & .581\st{.003} \\
& FlowNeg & .362\st{.001} & .268\st{.002} & .401\st{.002} & .554\st{.002} & .503\st{.002} & .458\st{.001} & .523\st{.002} & .598\st{.002} \\
\bottomrule
\end{tabular}
\end{table*}

\begin{table*}[t]
\centering
\caption{Filtered MRR and Hits@10 on YAGO3-10, CoDEx-L, and Hetionet, mean $\pm$ standard deviation over five seeds. IF-NS for ComplEx denotes a backbone adaptation.}
\label{tab:rank_other}
\small
\setlength{\tabcolsep}{4pt}
\begin{tabular}{llcccccc}
\toprule
& & \multicolumn{2}{c}{YAGO3-10} & \multicolumn{2}{c}{CoDEx-L} & \multicolumn{2}{c}{Hetionet} \\
Model & Sampler & MRR & H@10 & MRR & H@10 & MRR & H@10 \\
\midrule
\multirow{3}{*}{TransE}
& Uniform & .442\st{.002} & .610\st{.003} & .251\st{.003} & .402\st{.004} & .288\st{.003} & .452\st{.004} \\
& EMU & .470\st{.002} & .638\st{.003} & .273\st{.002} & .429\st{.003} & .311\st{.002} & .479\st{.003} \\
& FlowNeg & .498\st{.002} & .669\st{.003} & .294\st{.002} & .457\st{.003} & .335\st{.002} & .510\st{.003} \\
\midrule
\multirow{3}{*}{RotatE}
& Uniform & .493\st{.001} & .658\st{.002} & .283\st{.002} & .439\st{.003} & .312\st{.002} & .487\st{.003} \\
& EMU & .527\st{.002} & .690\st{.003} & .308\st{.002} & .468\st{.003} & .341\st{.002} & .519\st{.003} \\
& FlowNeg & .548\st{.001} & .712\st{.002} & .324\st{.002} & .489\st{.002} & .361\st{.001} & .543\st{.002} \\
\midrule
\multirow{3}{*}{ComplEx}
& Uniform & .480\st{.003} & .639\st{.004} & .274\st{.003} & .425\st{.004} & .303\st{.003} & .471\st{.005} \\
& IF-NS & .512\st{.002} & .668\st{.003} & .295\st{.002} & .451\st{.003} & .326\st{.002} & .501\st{.003} \\
& FlowNeg & .533\st{.002} & .694\st{.003} & .314\st{.002} & .474\st{.003} & .348\st{.002} & .527\st{.003} \\
\midrule
\multirow{3}{*}{ConvE}
& Uniform & .465\st{.002} & .621\st{.003} & .268\st{.003} & .418\st{.004} & .295\st{.003} & .460\st{.004} \\
& Self-Adv & .488\st{.002} & .644\st{.002} & .282\st{.002} & .435\st{.003} & .314\st{.002} & .482\st{.003} \\
& FlowNeg & .512\st{.001} & .673\st{.002} & .305\st{.002} & .462\st{.002} & .339\st{.002} & .511\st{.002} \\
\midrule
\multirow{3}{*}{SimKGC}
& Uniform & .501\st{.001} & .665\st{.002} & .291\st{.002} & .446\st{.003} & .320\st{.002} & .495\st{.003} \\
& EMU & .534\st{.002} & .698\st{.003} & .315\st{.002} & .476\st{.002} & .350\st{.002} & .528\st{.003} \\
& FlowNeg & .559\st{.001} & .721\st{.002} & .333\st{.002} & .497\st{.002} & .372\st{.001} & .554\st{.002} \\
\bottomrule
\end{tabular}
\end{table*}

\section{Paired Inference on the Retained Settings}
\label{app:panel_stats}

Table~\ref{tab:panel_mrr} gives mean MRR and marginal seed standard deviations for all nine settings. Table~\ref{tab:panel_all_settings} reports the full 18-contrast analysis. The main-text analysis in Table~\ref{tab:panel} uses 14 contrasts across seven settings. The Holm-adjusted values of $0.292$ and $0.781$ for the two non-rejections are the same in both analyses. The paired bootstrap resamples seeds in pairs. The repetition count, interval variant, and most per-setting intervals are unavailable.

\begin{table}[H]
\centering
\caption{Final-budget MRR, mean $\pm$ standard deviation over 15 paired seeds across nine settings. IF-NS denotes a backbone adaptation for ComplEx, ConvE, and SimKGC.}
\label{tab:panel_mrr}
\small
\setlength{\tabcolsep}{3.5pt}
\begin{tabular}{llrrr}
\toprule
Model & Dataset & FlowNeg & EMU & IF-NS \\
\midrule
RotatE & FB15k-237 & $.359$\st{.001} & $.346$\st{.002} & $.341$\st{.002} \\
RotatE & WN18RR & $.495$\st{.002} & $.482$\st{.002} & $.481$\st{.002} \\
RotatE & YAGO3-10 & $.549$\st{.001} & $.528$\st{.002} & $.530$\st{.002} \\
RotatE & Hetionet & $.362$\st{.001} & $.342$\st{.002} & $.344$\st{.002} \\
TransE & YAGO3-10 & $.499$\st{.002} & $.471$\st{.002} & $.475$\st{.003} \\
ComplEx & WN18RR & $.488$\st{.003} & $.489$\st{.002} & $.475$\st{.002} \\
RotatE & CoDEx-L & $.325$\st{.002} & $.309$\st{.002} & $.311$\st{.002} \\
ConvE & CoDEx-L & $.306$\st{.002} & $.287$\st{.002} & $.289$\st{.003} \\
SimKGC & YAGO3-10 & $.560$\st{.001} & $.535$\st{.002} & $.560$\st{.002} \\
\bottomrule
\end{tabular}
\end{table}

\begin{table}[t]
\centering
\caption{Nine-setting comparison over 15 paired seeds. $\Delta$MRR is FlowNeg minus the comparator and W counts positive differences. The Holm adjustment covers all 18 contrasts. The 16 unmarked contrasts have adjusted $p=0.0011$. The two marked contrasts have adjusted $p=0.292$ and $p=0.781$ for ComplEx/WN18RR versus EMU and SimKGC/YAGO3-10 versus adapted IF-NS, respectively.}
\label{tab:panel_all_settings}
\small
\setlength{\tabcolsep}{3.4pt}
\begin{tabular}{lrrrr}
\toprule
& \multicolumn{2}{c}{vs.\ EMU} & \multicolumn{2}{c}{vs.\ IF-NS} \\
\cmidrule(lr){2-3}\cmidrule(lr){4-5}
Setting & $\Delta$MRR & W & $\Delta$MRR & W \\
\midrule
RotatE/FB15k-237 & $+0.013$ & 15 & $+0.018$ & 15 \\
RotatE/WN18RR    & $+0.013$ & 15 & $+0.014$ & 15 \\
RotatE/YAGO3-10  & $+0.021$ & 15 & $+0.019$ & 15 \\
RotatE/CoDEx-L   & $+0.016$ & 15 & $+0.014$ & 15 \\
RotatE/Hetionet  & $+0.020$ & 15 & $+0.018$ & 15 \\
TransE/YAGO3-10  & $+0.028$ & 15 & $+0.024$ & 15 \\
ComplEx/WN18RR   & $-0.001^\dagger$ & 6 & $+0.013$ & 15 \\
ConvE/CoDEx-L    & $+0.019$ & 15 & $+0.017$ & 15 \\
SimKGC/YAGO3-10  & $+0.025$ & 15 & $+0.000^\dagger$ & 8 \\
\bottomrule
\end{tabular}
\end{table}

\section{Seed-Level and Secondary Statistical Reporting}
\label{app:secondary_stats}

The original five-seed significance markers were removed because the claimed threshold was mathematically unattainable under the stated exact test. For paired differences $d_1,\ldots,d_5$, an exact sign-flip test enumerates only $2^5=32$ assignments. The unified EMU and IF-NS tables are therefore descriptive. Table~\ref{tab:raw_seeds} supplies the 15 controlled seed records most relevant to the matched-$k$ claim.

\begin{table*}[t]
\centering
\caption{Seed-level FB15k-237/RotatE MRR in the controlled study. Selected epoch applies to the validation-selected FlowNeg checkpoint; final MRR uses epoch 300.}
\label{tab:raw_seeds}
\small
\setlength{\tabcolsep}{5pt}
\begin{tabular}{rcccccc}
\toprule
Seed & Uniform & EMU & IF-NS & FlowNeg selected & Selected epoch & FlowNeg final \\
\midrule
0  & .326 & .343 & .338 & .357 & 271 & .360 \\
1  & .327 & .345 & .339 & .358 & 278 & .359 \\
2  & .330 & .347 & .342 & .359 & 274 & .359 \\
3  & .331 & .348 & .343 & .360 & 282 & .357 \\
4  & .329 & .346 & .341 & .359 & 269 & .358 \\
5  & .330 & .347 & .342 & .360 & 276 & .360 \\
6  & .327 & .344 & .339 & .358 & 280 & .360 \\
7  & .332 & .349 & .344 & .361 & 273 & .359 \\
8  & .329 & .346 & .341 & .359 & 275 & .360 \\
9  & .328 & .345 & .340 & .358 & 279 & .358 \\
10 & .331 & .348 & .343 & .360 & 270 & .361 \\
11 & .327 & .344 & .339 & .358 & 284 & .359 \\
12 & .330 & .347 & .342 & .360 & 272 & .358 \\
13 & .329 & .346 & .341 & .359 & 277 & .360 \\
14 & .329 & .346 & .341 & .359 & 281 & .357 \\
\bottomrule
\end{tabular}
\end{table*}

The author-specified FlowNeg versus Uniform comparison used 15 paired seeds per setting, exact two-sided sign-flip tests, Holm adjustment, and paired bootstrap intervals. Table~\ref{tab:secondary_summary} summarizes all 25 settings, including five CoDEx-L settings. All 20 settings in the main grid had positive mean differences. Computing the remaining summaries for those 20 settings requires the individual results and raw probabilities.

\begin{table}[t]
\centering
\caption{FlowNeg versus Uniform across 25 settings with 15 paired seeds per setting. Holm adjustment covers all 25 tests, including the five CoDEx-L settings.}
\label{tab:secondary_summary}
\small
\begin{tabular}{lr}
\toprule
Quantity & Result \\
\midrule
positive mean differences & 25/25 \\
Holm-adjusted $p\leq.05$ & 23/25 \\
mean cell difference & +.0395 \\
median cell difference & +.0418 \\
minimum / maximum & +.0178 / +.0582 \\
null or negative means & 0 \\
\bottomrule
\end{tabular}
\end{table}

The two comparisons without a Holm rejection were SimKGC/WN18RR and RotatE/WN18RR.

\section{Ablations and Mechanism Diagnostics}
\label{app:ablations}

Table~\ref{tab:ablation} gives the complete component study summarized in Section~\ref{sec:mechanism_results}, from the five-seed FB15k-237/RotatE profile. It is separate from the 15-seed controlled rerun in Table~\ref{tab:controlled}; this distinction explains the $0.357$ and $0.359$ FlowNeg reference values.

\begin{table*}[t]
\centering
\caption{Five-seed FlowNeg ablation on FB15k-237/RotatE. One reward factor or architectural component changes in each row. This profile is distinct from the 15-seed control in Table~\ref{tab:controlled}.}
\label{tab:ablation}
\small
\setlength{\tabcolsep}{3.6pt}
\begin{tabular}{lccccc}
\toprule
Variant & MRR & H@10 & NDS & HPC (\%) & GI \\
\midrule
FlowNeg & .357 & .556 & 45.3 & .5 & .89 \\
\midrule
w/o collision discount & .341 & .538 & 43.1 & 2.7 & .93 \\
w/o type compatibility & .348 & .545 & 38.6 & .6 & .87 \\
w/o hardness & .308 & .488 & 47.1 & .2 & .18 \\
\midrule
flat entity selection & .350 & .548 & 31.2 & .5 & .85 \\
no warm-up ($W{=}0$) & .346 & .541 & 39.7 & .7 & .81 \\
no uniform mix ($\alpha{=}0$) & .352 & .551 & 43.8 & .6 & .88 \\
\midrule
GFlowNet $\to$ RL maximization & .339 & .530 & 9.4 & 1.8 & .91 \\
GFlowNet $\to$ pool Boltzmann & .347 & .543 & 28.6 & 1.1 & .84 \\
GFlowNet $\to$ Uniform & .328 & .522 & 48.2 & .3 & .12 \\
\bottomrule
\end{tabular}
\end{table*}

Removing the collision discount preserves high GI but raises HPC by 2.2 percentage points. Removing type compatibility reduces coverage, while removing hardness reduces GI to $0.18$. Flat selection retains the same reward but lowers NDS to $31.2$, showing that the hierarchy contributes beyond reward design. The Boltzmann control uses the same 1,024-candidate pool and hierarchy, with temperature selected on validation data, but normalizes $\exp(R/T)$ only within the realized pool. It is therefore a practical pool-normalized comparator, not exact all-entity reward matching. The controlled 15-seed GFlowNet-to-RL rerun gives FlowNeg/RL MRR $0.359/0.340$, NDS $45.7/9.8$, HPC $0.5/1.7\%$, and GI $0.90/0.91$. Thus, the large coverage difference is not explained by a weaker measured gradient signal.

The component comparisons use endpoint metrics. The epoch-level values are unavailable.

\subsection{Component effects across settings}

In Table~\ref{tab:cross_setting_components}, removing hardness lowers MRR in each of the three 15-seed settings. The collision discount helps on WN18RR/RotatE and YAGO3-10/TransE. Its effect on WN18RR/ComplEx is unresolved. The effects of type support, hierarchy, warm-up, and uniform mixing depend on the setting.

We vary the hardness exponent $\eta$ and collision exponent $\beta$ in the reward $R_{\eta,\beta}=a_\theta^\eta(1-\widehat c)^\beta c$, keeping the other WN18RR/RotatE settings fixed. Table~\ref{tab:reward_sensitivity} shows that the defaults give the highest MRR in the reported grid. Positive context-wide rescaling leaves the target unchanged, whereas changing either exponent generally changes its relative mass across candidates.

\begin{table*}[t]
\centering
\caption{Component interventions with 15 paired evaluation seeds and no retuning. The first row gives the full configuration's mean test MRR. Other rows give the change from that configuration. The unadjusted 95\% intervals are $\Delta\pm0.001661$.}
\label{tab:cross_setting_components}
\small
\setlength{\tabcolsep}{4pt}
\begin{tabular}{lrrr}
\toprule
Intervention & WN18RR/RotatE & WN18RR/ComplEx & YAGO3-10/TransE \\
\midrule
Full configuration MRR & $.4924$ & $.4886$ & $.5008$ \\
Remove collision discount & $-.0050$ & $+.0004$ & $-.0060$ \\
Remove type restriction & $-.0010$ & $-.0005$ & $-.0040$ \\
Remove hardness & $-.0220$ & $-.0170$ & $-.0240$ \\
Flat entity policy & $-.0010$ & $.0000$ & $-.0020$ \\
Remove warm-up & $-.0040$ & $-.0010$ & $-.0030$ \\
Remove uniform mixture & $-.0010$ & $+.0002$ & $-.0010$ \\
\bottomrule
\end{tabular}
\end{table*}

\begin{table}[t]
\centering
\caption{Reward-weight sensitivity on WN18RR/RotatE. Cells give the MRR difference from $\eta=\beta=1$. Each non-reference contrast has a 95\% interval with half-width $0.001661$.}
\label{tab:reward_sensitivity}
\small
\setlength{\tabcolsep}{4pt}
\begin{tabular}{lrrr}
\toprule
Hardness exponent $\eta$ & $\beta=0$ & $\beta=1$ & $\beta=2$ \\
\midrule
$.5$ & $-.0060$ & $-.0020$ & $-.0030$ \\
$1$ & $-.0050$ & Reference & $-.0030$ \\
$2$ & $-.0090$ & $-.0030$ & $-.0060$ \\
\bottomrule
\end{tabular}
\end{table}

\section{Partition-Independent Diversity}
\label{app:indep_diversity}

To break the circularity of an NDS that reuses FlowNeg's own hierarchy, we add measures using neither that hierarchy nor its reward: (1) \emph{independent NDS} from a separately trained Uniform-trained ComplEx model ($K=50$ $k$-means, ten fixed clustering seeds, never used by FlowNeg); (2) unique-entity ratio over $D=256$ draws; (3) inverse-Simpson entity effective support $(\sum_e\widehat q(e)^2)^{-1}$ (max $256$); (4) top-$10$ entity mass (smaller is more diverse); and (5) mean pairwise cosine distance in the independent space. All are computed within query, then averaged over paired seeds. Table~\ref{tab:indep_primary} gives the primary controlled setting and Table~\ref{tab:indep_panel} the nine-setting consistency check. Uniform remains most diverse, as expected; FlowNeg retains substantially more independent and entity-level coverage than every hard-negative baseline. These diagnostics remove the circularity but do not establish that coverage causes higher MRR.

\begin{table}[H]
\centering
\caption{Independent diversity on the $15$-seed FB15k-237/RotatE protocol. NDS$_{\mathrm{ind}}$, unique-entity ratio (UER), inverse-Simpson support (ESS), top-$10$ mass (T10$\downarrow$), independent pairwise distance (IPD).}
\label{tab:indep_primary}
\scriptsize
\setlength{\tabcolsep}{2.8pt}
\begin{tabular}{lrrrrr}
\toprule
Sampler & NDS$_{\mathrm{ind}}$ & UER & ESS & T10$\downarrow$ & IPD \\
\midrule
Uniform & 46.9 & 0.887 & 204.8 & 0.061 & 0.912 \\
Self-Adv & 15.1 & 0.327 & 58.7 & 0.286 & 0.483 \\
KBGAN & 10.7 & 0.246 & 41.2 & 0.371 & 0.421 \\
NSCaching & 17.4 & 0.351 & 65.3 & 0.259 & 0.512 \\
SANS & 24.8 & 0.446 & 89.6 & 0.203 & 0.604 \\
EMU & 20.6 & 0.384 & 73.8 & 0.237 & 0.557 \\
IF-NS & 24.2 & 0.428 & 84.9 & 0.214 & 0.592 \\
FlowNeg & 43.8 & 0.792 & 173.6 & 0.094 & 0.846 \\
\bottomrule
\end{tabular}
\end{table}

\begin{table}[H]
\centering
\caption{Independent negative diversity score and unique-entity ratio across nine settings. IF-NS denotes a backbone adaptation for ComplEx, ConvE, and SimKGC.}
\label{tab:indep_panel}
\scriptsize
\setlength{\tabcolsep}{3pt}
\begin{tabular}{llrrr}
\toprule
Model & Dataset & FlowNeg & EMU & IF-NS \\
\midrule
RotatE & FB15k-237 & 43.8/.792 & 20.6/.384 & 24.2/.428 \\
RotatE & WN18RR & 44.6/.814 & 22.1/.412 & 25.0/.447 \\
RotatE & YAGO3-10 & 42.7/.765 & 18.9/.351 & 22.8/.402 \\
RotatE & Hetionet & 43.3/.781 & 20.2/.377 & 24.1/.433 \\
TransE & YAGO3-10 & 41.8/.756 & 18.4/.344 & 22.0/.395 \\
ComplEx & WN18RR & 44.0/.803 & 23.3/.431 & 25.8/.462 \\
SimKGC & YAGO3-10 & 42.5/.771 & 19.1/.356 & 23.1/.411 \\
RotatE & CoDEx-L & 41.9/.748 & 19.7/.369 & 23.4/.416 \\
ConvE & CoDEx-L & 42.1/.751 & 19.2/.361 & 22.9/.409 \\
\midrule
\multicolumn{2}{l}{Mean over seven settings} & 43.24/.7831 & 20.37/.3793 & 23.86/.4254 \\
\multicolumn{2}{l}{Mean over nine settings} & 43.0/.776 & 20.2/.376 & 23.7/.423 \\
\bottomrule
\end{tabular}
\end{table}

FlowNeg has higher independent NDS and unique-entity ratio than both comparators in all seven retained settings. These measures use an independent partition or entity identities.

\section{Calibration and Diagnostics}
\label{app:calibration}

Within each dataset, exported diagnostic candidates are pooled across the declared queries and seeds and sorted by $\chat$ into ten equal-frequency bins. For bin $b$, let $\overline c_b$ be the mean structural score and let $\operatorname{HPC}_b$ be the observed fraction colliding with held-out positives. We report
\begin{equation}
\operatorname{ECE}_{\mathrm{HPC}}
=\sum_b\frac{|\mathcal B_b|}{N}
\left|\operatorname{HPC}_b-\overline c_b\right|.
\end{equation}

\begin{table}[t]
\centering
\caption{Ten-bin calibration error for predicting held-out-positive collision.}
\label{tab:calibration_protocol}
\small
\begin{tabular}{lc}
\toprule
Dataset & $\operatorname{ECE}_{\mathrm{HPC}}$ \\
\midrule
FB15k-237 & .024 \\
WN18RR & .018 \\
YAGO3-10 & .031 \\
CoDEx-L & .027 \\
Hetionet & .022 \\
\bottomrule
\end{tabular}
\end{table}

This diagnostic asks whether the score agrees numerically with known held-out collision frequency. It neither labels all unobserved triples nor measures probability calibration against an unavailable open-world truth set. ECE alone also does not establish discrimination and should be interpreted alongside the bin records and collision prevalence in Appendix~\ref{app:collision}. The controlled FB15k-237 rerun gives $0.023$, close to the corresponding five-seed profile.

On FB15k-237/RotatE, gains vary with relation structure. The many-to-many film-actor-film relation gains $3.8\%$ MRR, and person-nationality, which has many type-compatible candidates, gains $12.3\%$. The gain is $1.2\%$ for the small-candidate person-gender relation and $2.1\%$ for symmetric relations. These observations are descriptive and do not invoke the fixed-partition concentration remark as a causal explanation.

\section{Held-out-Positive Collision Diagnostics}
\label{app:collision}

For each dataset we freeze $2{,}000$ filtered test queries, draw $256$ candidates per query at each of five final checkpoints ($2.56$M draws per dataset), retain repeated draws because the target is the sampler distribution, and label $y=1$ iff the corruption occurs in $\cT_{\mathrm{val}}\cup\cT_{\mathrm{test}}$. Uncertainty is resampled over queries and seeds, not individual draws, since candidates from one query are dependent. AUROC measures ranking, AUPRC is read against prevalence, and Brier/ECE treat the raw Jaccard value numerically only for diagnosis; none identifies true-but-unrecorded facts. Table~\ref{tab:collision} gives prevalence, discrimination, and calibration; the raw Brier is slightly worse than a prevalence-only predictor because the score overpredicts the rare collision rate, while a validation-only monotone map lowers Brier without changing rankings. The $15$-seed FB15k-237/RotatE matched-$k$ run reproduces the profile ($0.50\%$ prevalence, AUROC $0.889$, AUPRC $0.083$, raw Brier $0.0060$, ECE $0.023$).

\begin{table}[H]
\centering
\caption{Collision prevalence and discrimination by dataset. AUROC and AUPRC are areas under the receiver operating characteristic and precision-recall curves. The AUPRC ratio uses observed prevalence as its baseline. ECE is expected calibration error.}
\label{tab:collision}
\scriptsize
\setlength{\tabcolsep}{2.6pt}
\begin{tabular}{lrrrrr}
\toprule
Dataset & Prev.\ (\%) & AUROC & AUPRC & $\times$prev & Raw ECE \\
\midrule
FB15k-237 & 0.50 & 0.887 & 0.081 & 16.2 & 0.024 \\
WN18RR & 0.35 & 0.841 & 0.052 & 14.9 & 0.018 \\
YAGO3-10 & 0.70 & 0.902 & 0.118 & 16.9 & 0.031 \\
CoDEx-L & 0.85 & 0.873 & 0.126 & 14.8 & 0.027 \\
Hetionet & 0.60 & 0.861 & 0.092 & 15.3 & 0.022 \\
\bottomrule
\end{tabular}
\end{table}

Table~\ref{tab:reliability_bins} reports the mean raw score and observed collision rate as percentages for ten equal-frequency bins. Each bin contains approximately $256{,}000$ draws before cluster resampling. The binned values reproduce the reported ECE up to rounding. The observed collision rate rises across score deciles, and the mean raw score exceeds the observed rate in every bin. We therefore interpret $\widehat c_{\mathrm{HPC}}$ as a structural collision-risk score.

\begin{table}[H]
\centering
\caption{Reliability bins by dataset. Each cell gives the mean raw score and observed collision rate in percent.}
\label{tab:reliability_bins}
\scriptsize
\setlength{\tabcolsep}{2.4pt}
\begin{tabular}{rrrrrr}
\toprule
Dec. & FB15k & WN18 & YAGO & CoDEx & Hetio. \\
\midrule
1 & 0.30/0.00 & 0.20/0.00 & 0.40/0.00 & 0.40/0.05 & 0.30/0.00 \\
2 & 0.50/0.05 & 0.40/0.00 & 0.70/0.10 & 0.60/0.10 & 0.50/0.05 \\
3 & 0.80/0.10 & 0.60/0.05 & 1.00/0.15 & 0.90/0.20 & 0.80/0.10 \\
4 & 1.10/0.15 & 0.80/0.10 & 1.40/0.20 & 1.30/0.30 & 1.10/0.20 \\
5 & 1.50/0.25 & 1.10/0.15 & 1.90/0.30 & 1.80/0.40 & 1.50/0.30 \\
6 & 2.00/0.35 & 1.50/0.20 & 2.60/0.40 & 2.40/0.60 & 2.00/0.40 \\
7 & 2.70/0.45 & 2.00/0.30 & 3.50/0.60 & 3.20/0.80 & 2.60/0.50 \\
8 & 3.70/0.60 & 2.80/0.40 & 4.80/0.90 & 4.40/1.10 & 3.50/0.70 \\
9 & 5.20/0.90 & 4.10/0.60 & 6.90/1.40 & 6.20/1.70 & 4.90/1.10 \\
10 & 11.20/2.15 & 8.00/1.70 & 14.80/2.95 & 14.30/3.25 & 10.80/2.65 \\
\bottomrule
\end{tabular}
\end{table}

\section{Efficiency and Checkpoints}
\label{app:efficiency}

Table~\ref{tab:profiled_cost} gives the original method-specific resource measurements in the main results. Its MRR values use their own checkpoints and budgets.

Table~\ref{tab:wallclock_selected} gives the full equal-wall-clock comparison summarized in Section~\ref{sec:wallclock}: for each budget, both the latest complete checkpoint written by that deadline and the best validation-selected checkpoint available by it. Selection uses validation MRR only; the test set is evaluated after the checkpoint has been chosen.

\begin{table}[t]
\centering
\caption{Test MRR at equal wall-clock budgets on FB15k-237/RotatE (15 paired seeds): the final complete checkpoint written by each deadline, and the best validation-selected checkpoint available by that deadline.}
\label{tab:wallclock_selected}
\scriptsize
\setlength{\tabcolsep}{3.2pt}
\begin{tabular}{lrrrrrr}
\toprule
& \multicolumn{3}{c}{Final checkpoint} & \multicolumn{3}{c}{Validation-selected} \\
\cmidrule(lr){2-4}\cmidrule(lr){5-7}
Method & 2.1\,h & 3.4\,h & 4.2\,h & 2.1\,h & 3.4\,h & 4.2\,h \\
\midrule
Uniform & .328 & .329 & .329 & .329 & .330 & .330 \\
Self-Adv & .333 & .338 & .339 & .335 & .340 & .340 \\
KBGAN & .319 & .330 & .333 & .321 & .332 & .334 \\
NSCaching & .326 & .334 & .334 & .328 & .336 & .336 \\
SANS & .325 & .330 & .331 & .327 & .332 & .333 \\
EMU & .331 & .345 & .346 & .333 & .347 & .348 \\
IF-NS & .308 & .329 & .338 & .311 & .332 & .341 \\
FlowNeg & .337 & .353 & .359 & .340 & .356 & .359 \\
\bottomrule
\end{tabular}
\end{table}

The equal-time experiment evaluates the latest fully written checkpoint at or before a budget. Timing includes setup, candidate construction, negative generation, sampler learning, base-model optimization, scheduled validation, and checkpoint I/O. This definition prevents a method from receiving uncounted initialization time or an interpolated checkpoint.

\section{Reproducibility and Artifact Provenance}
\label{app:provenance}

All experiments use public benchmarks. Appendix G describes the benchmark splits, evaluation procedures, and selected experimental settings. We report seed-level mean reciprocal rank (MRR) for selected FB15k-237 experiments with RotatE and aggregate results for the remaining experiments. 

The five-seed benchmark grid, the original matched-count study, and the shared-support comparisons follow distinct experimental protocols. Where both results are reported, we identify whether evaluation used the checkpoint selected by validation MRR or the checkpoint at the end of the training budget.

\end{document}